\documentclass[twoside]{article}
\usepackage[accepted]{aistats2025}
\usepackage{amsthm,amsmath,amssymb,amsbsy,bbm,mathrsfs,supertabular,eurosym,graphicx,enumitem}
\usepackage{bm}
\usepackage{xcolor}
\usepackage{mathtools}
\newtheoremstyle{BBstyle0}  {}{}{\itshape}{}{\bfseries}{}{6pt}{}
\newtheoremstyle{BBstyle1}  {3pt}{3pt}{\rmfamily}{}{\itshape}{: }{3pt}{}
\newtheoremstyle{BBstyle2}  {3pt}{3pt}{\itshape}{}{\bfseries\large}{}{0pt}{}
\newtheoremstyle{BBstyle3}  {}{}{\itshape}{}{\bfseries}{: }{3pt}{}
\newtheoremstyle{BBstyle4}  {}{}{\rmfamily}{}{\bfseries}{}{6pt}{}
\usepackage[normalem]{ulem}
\usepackage{subcaption}
\usepackage{graphicx}
\definecolor{jcgreen}{rgb}{0.09, 0.65, 0.35}

\usepackage[utf8]{inputenc}
\usepackage[T1]{fontenc}
\usepackage{yhmath}

\usepackage{amsthm,amsmath,amssymb,amsbsy,bbm,mathrsfs,supertabular,
eurosym,graphicx,enumitem,xcolor}
\usepackage{mathtools}

\newtheoremstyle{BBstyle0}  {}{}{\itshape}{}{\bfseries}{}{6pt}{}
\newtheoremstyle{BBstyle1}  {3pt}{3pt}{\rmfamily}{}{\itshape}{: }{3pt}{}
\newtheoremstyle{BBstyle2}  {3pt}{3pt}{\itshape}{}{\bfseries\large}{}{0pt}{}
\newtheoremstyle{BBstyle3}  {}{}{\itshape}{}{\bfseries}{: }{3pt}{}
\newtheoremstyle{BBstyle4}  {}{}{\rmfamily}{}{\bfseries}{}{6pt}{}
  
\usepackage[normalem]{ulem} 
\newtheorem{thm}{Theorem}

\newtheorem{prop}{Proposition}

\newtheorem{ass}{Assumption}

\theoremstyle{definition}

\usepackage[english]{babel}

\newcommand{\cro}[1]{\left[{#1}\right]}

\newcommand{\argmin}{\mathop{\rm argmin}}

\newcommand{\Var}{\mathop{\rm Var}\nolimits}

\newcommand{\Tr}{\mathop{\rm Tr}\nolimits}

\newcommand{\E}{{\mathbb{E}}}

\newcommand{\R}{{\mathbb{R}}}

\newcommand{\Z}{{\mathbb{Z}}}

\DeclareMathAlphabet{\mathscrbf}{OMS}{mdugm}{b}{n}

\newcommand{\cE}{{\mathcal{E}}}
\newcommand{\cF}{{\mathcal{F}}}

\newcommand{\cN}{{\mathcal{N}}}

\newcommand{\bs}[1]{\boldsymbol{#1}}

\newlist{lista}{enumerate}{1}
\setlist[lista,1]{label=\alph*),ref=\alph*)}

\newlist{listi}{enumerate}{1}
\setlist[listi,1]{label=(\roman*),ref=(\roman*),align=left}

\newcommand{\eref}[1]{(\ref{#1})}

\newcommand{\eps}{{\varepsilon}}

\newcommand{\id}{\operatorname{id}}

\newcommand{\Bias}{\operatorname{Bias}}
\usepackage{natbib}

\renewcommand{\bibsection}{\subsubsection*{\bibname}}
\RequirePackage[colorlinks,citecolor=blue,urlcolor=blue]{hyperref}
\begin{document}
\runningauthor{Chen, Schmidt-Hieber, Donnat, Klopp}

\twocolumn[
\aistatstitle{Understanding the Effect of GCN Convolutions in Regression Tasks}

\aistatsauthor{ Juntong Chen \And Johannes Schmidt-Hieber\And Claire Donnat\And Olga Klopp}
\aistatsaddress{University of Twente\And University of Twente\And University of Chicago\And ESSEC and ENSAE} 
]

\begin{abstract}
Graph Convolutional Networks (GCNs) have become a pivotal method in machine learning for modeling functions over graphs. Despite their widespread success across various applications, their statistical properties (e.g., consistency, convergence rates) remain ill-characterized. To begin addressing this knowledge gap, we consider networks for which the graph structure implies that neighboring nodes exhibit similar signals and provide statistical theory for the impact of convolution operators. Focusing on estimators based solely on neighborhood aggregation, we examine how two common convolutions---the original GCN and GraphSAGE convolutions---affect the learning error as a function of the neighborhood topology and the number of convolutional layers.  We explicitly characterize the bias-variance type trade-off incurred by GCNs as a function of the neighborhood size and identify specific graph topologies where convolution operators are less effective. Our theoretical findings are corroborated by synthetic experiments, and provide a start to a deeper quantitative understanding of convolutional effects in GCNs for offering rigorous guidelines for practitioners.
\end{abstract}

\section{INTRODUCTION}
Graph Convolutional Networks (GCNs) have become one of the preferred tools for modeling, analyzing, and predicting signals on graphs \citep{hamilton2017inductive,Max}. Despite their impressive success on academic benchmarks, several fundamental issues limit their broader applicability and reliability in real-world scenarios. In particular, while graphs are highly diverse in their properties (e.g., sparsity, degree distribution, node feature types) and the relationships they encode, GCNs are often suggested as a one-size-fits-all approach. This leaves practitioners with the challenge of selecting an appropriate convolution and architecture for their specific task. Without a clear understanding of the inductive biases these convolutions encode and their expected performance in relation to dataset properties, choosing among the sixty-six available convolution options in pytorch geometric can become a daunting task \citep{pytorch2019}. To bridge this knowledge gap, this paper examines the properties of graph convolutions and their relationship to the graph's neighborhood characteristics, with a particular focus on regression tasks over networks.

\paragraph{Statistical Setting.}  We consider the problem of network regression under fixed design. Let $\mathcal{G} = (\mathcal{V},\mathcal{E})$  denote a graph with vertex set  $\mathcal{V}$ and edge set $\mathcal{E}$. Without loss of generality, we index the vertices from 1 to $n$, so $\mathcal{V} = \{1, \ldots, n\}$ and $|\mathcal{V}| = n$. Assume that at each node $i\in\{1,\ldots,n\}$, we observe a corresponding real-valued response variable $Y_i$, which is generated according to the following mechanism:
\begin{equation}\label{eq:model}
    Y_i=f_i^{*}+\eps_i,
\end{equation} where the noise variables $\eps_i$ are assumed to be uncorrelated, centered, and have variance equal to one. Let $\mathbf{Y} := (Y_{1}, \ldots, Y_n)^{\top} \in \mathbb{R}^{n}$ denote the vector of responses from all nodes in the graph $\mathcal{G}$, ${\bm{\eps}}:=(\eps_1,\ldots,\eps_n)^{\top}\in \mathbb{R}^{n}$ the noise vector, and ${\mathbf{f^{*}:}=(f_1^{*},\ldots,f_n^{*})^{\top}\in \mathbb{R}^{n}}$ the regression vector, which we also refer to as the signals. Throughout the paper, we assume that neighboring nodes in $\mathcal{G}$ admit similar signals: $f_i^* \approx f_j^*$, if $(i,j) \in \mathcal{E}$. Our goal is to recover $\mathbf{f^{*}}$ from the observations $\mathbf{Y}$. 

To reconstruct $\mathbf{f^{*}}$, a possible solution is to minimize the least squares objective:
\begin{align*}
\mathbf{\hat f}_{\cF}:=\argmin_{g\in\cF}\big\|\mathbf{Y}-g(\mathbf{Y})\big\|_2^2
\end{align*}
with $\cF$ being a prespecified function class. To incorporate information about the graph, graph-based regularization methods consider instead a penalized objective of the form: 
\begin{align*}
\mathbf{\hat f}_{\lambda}  
:=&\argmin_{g\in\cF}\big\|\mathbf{Y}-g(\mathbf{Y})\big\|_2^2\nonumber\\
&+\lambda 
\sum_{(i,j)\in \cE}\text{Pen} \big(g(\mathbf{Y})_i-g(\mathbf{Y})_j\big) 
\end{align*}
where $\text{Pen}$ denotes a function penalizing differences along edges of the graph, and $g(\mathbf{Y})_i$ represents the $i$-th component of $g(\mathbf{Y})\in \mathbb{R}^{n}$.

When the graph is endowed with a node feature matrix $\mathbf{X} \in \mathbb{R}^{n \times p}$, and the estimator $\mathbf{\hat{f}}$ is chosen to be a function of these features (e.g. $\mathbf{\hat{f}} =\mathbf{\hat{f}}(\mathbf{X})$), the task is usually known as {\it node regression}. When the estimator is constructed using information solely from the observed values of $\mathbf{Y}\in\mathbb{R}^n$, the problem boils down to a denoising task usually known as \textit{graph-trend filtering} \citep{hutter2016optimal,wang2016trend}, an extension of the classical Gaussian sequence model \citep{rigollet201518,wasserman2006all} to accommodate graph structure. In this paper, we aim to study the performance of graph convolutions in providing reliable estimates for $\mathbf{f^{*}}$ under the denoising setting. Specifically, the feature $X_i$ associated with each node $i$ is simply given by $X_i=i$.

\paragraph{Contributions.} The main contributions of this paper are summarized as follows:
\begin{itemize}[noitemsep]
\item We establish a bias-variance type trade-off for the resulting denoising estimator in terms of the number of convolutional layers. This trade-off highlights how increasing the number of layers can reduce variance while potentially introducing greater bias, thereby affecting the overall estimation accuracy.
\item By showing that the variance of the GCN is a weighted sum over paths, we relate the variance of GCNs to that of the uniformly local averaging estimator and introduce the novel walk analysis approach to investigate the problem. This method may pave the way for further statistical studies of GCNs.
\item Based on the proposed walk analysis, we further quantitatively demonstrate the different variance decay behaviors under distinct local topologies of the graph. Specifically, we show that when high edge degree nodes are connected to low edge degree nodes this can dominate the variance and result in much slower overall variance decay.
\item Additionally, we provide a concrete example of the graph structure to illustrate the over-smoothing phenomenon, which serves as a prototype to explain why simply stacking layers leads to poor performance in practice. Whereas previous work has provided evidence of this phenomenon for asymptotically large graphs and classification settings \citep{oono2019graph,cai2020}, we identified a non-asymptotic regression setting.
\end{itemize}

\section{RELATED WORK}
The theory for node regression has been derived from different viewpoints. Asymptotic viewpoint assumes that the number of vertices $n$ tends to infinity, so that the Laplacian matrix is used to approximate the Laplace-Beltrami operator on an underlying manifold \citep{belkin2003laplacian,lafon2004diffusion,singer2006graph} on which the function $\mathbf{f^*}$ is assumed to be smooth \citep{hein2006uniform,von2008consistency}. Instead of assuming that the number of vertices $n$ tends to infinity, one can alternatively work with a fixed graph where, for a randomly selected subset of nodes, the response vectors are masked/unobserved \citep{Max, donnat2024one}. A typical assumption consists in assuming that neighboring nodes will have the same response. In this setting, several strategies leveraging the graph structure to guide the inference can be deployed: (a) Laplacian regularization, (b) $\ell_1$ regularization and (c) graph convolutions.

\textit{Laplacian regularization} (also known as Laplacian smoothing \citep{smola2003kernels}, a special instance of Tikhonov regularization) is perhaps one of the better studied version of the problem. As described in a previous paragraph, these methods usually consider a smooth function (where the smoothness of the function $\mathbf{f^{*}}$ is consequently defined via the eigenvalue decay of the graph Laplacian), and use the convergence of the graph Laplacian to the Laplace-Beltrami operator to establish the consistency and convergence rates of the proposed estimator  \citep{hein2006uniform,von2008consistency}. Several research groups have explored similar concepts.  \cite{belkin2004regularization} provides bounds on the generalization error for Tikhonov regularization and interpolated regularization.
Work by \cite{Kirichenko} derives posterior contraction rates for $\mathbf{f^{*}}$ using a Bayesian formalism. The rates depend on the eigenvalue decay of the graph Laplacian and a smoothness index that is defined via a Sobolev-type space based on the graph Laplacian.  More recently, work by Green and coauthors \citep{Green,Green2} derives optimal convergence rates for Laplacian-based regularization in the fixed design setting.

\textit{Graph-trend filtering} (also known as ``graph total-variation'' \citep{hutter2016optimal} or ``fused lasso'' \citep{tibshirani2005sparsity,tibshirani2011solution}) estimators use an $\ell_1$ penalty to guide the inference, so that estimators are derived as the solution of the following optimization problem:
$$  \mathbf{\hat{f}} \in \text{argmin}_{ \mathbf{x} \in \mathbb{R}^n} \| \mathbf{Y} - \mathbf{x}\|_2^2  + \lambda \sum_{(i,j)\in\mathcal{E}} | x_i -x_j|.$$
Recent work (e.g., see \cite{donnat2024one,hutter2016optimal, padilla_dfs_2018}) has characterized the convergence rate of these estimators as a function of the underlying graph smoothness, defined by the amount of variation along the graph edges: $\sum_{(i,j)\in\mathcal{E}} | f^*_i - f^*_j|.$ While this approach is attractive for approximating near piecewise-constant functions over graphs, the estimators are often more computationally intensive than Laplacian-based estimators.

Finally, {\it Graph convolutional networks} \citep{Max} are a more recent addition to the set of methods. While typically used for classification, this method is increasingly used for node regression tasks. The theoretical understanding of GCN models remains quite limited. For example, \cite{verma2019stability} analyzes the stability of single-layer GCN models and derives their generalization guarantees, showing that, in regression problems, the generalization gap between training and testing errors decreases at a sub-linear rate as the number of training samples increases. The work in \cite{lecun} and \cite{Defferrard} demonstrates that GCNs can be interpreted as a simplification of spectral-type Graph Neural Networks (GNNs) that makes use of the graph Laplacian. In \cite{li2018deeper}, it is shown that graph convolution is essentially a form of Laplacian smoothing. \cite{poly-GNN} investigates the classification performance of GNNs with graph-polynomial features under a general contextual stochastic block model. 

Contrary to earlier work, in this paper we aim to provide statistical insight into the regression setting. We analyze a widely used class of GNNs, namely GCNs, which we discuss in detail in the next section.
\paragraph{Notation.}
Let $A=(A_{ij})_{i, j\in \mathcal{V}}\in \R^{n \times n}$ denote the binary adjacency matrix, which encodes the edge structure of the graph $\mathcal{G}$ as follows:
\begin{align*}
A_{ij}:=
\begin{cases}
    1, &\text{if $i$ and $j$ are connected by an edge,} \\
0, &\text{otherwise}.    
\end{cases}
\end{align*}
For any $k=1,2,\ldots$, let $\mathcal{N}^k(i)$ represent the nodes within $k$ hops of node $i$. The neighborhood of node $i$ is denoted by $\mathcal{N}(i)$, with $\mathcal{N}^1(i)=\mathcal{N}(i)\cup\{i\}$. For $i\in\mathcal{V}$, let $d_i:=|\mathcal{N}(i)|$ be the edge degree of node $i$. The degree matrix $D$ is the $n\times n$ diagonal matrix with diagonal entries $D_{ii}=d_i,$ $i=1,\ldots,n.$ For a vector ${\bf{v}}=(v_1,\ldots,v_n)\in\R^{n}$, $\|{\bf{v}}\|_{2}^2=\sum_{i=1}^{n}v_i^2$ denotes the squared Euclidean norm and the notation $({\bf{v}})_{i}=v_i$ represents the $i$-th component of the vector. For a square matrix $M$, let $\|M\|_{F}$ denote the Frobenius norm and $\Tr(M)$ denote the trace. 

\section{GRAPH CONVOLUTIONAL NETWORKS}\label{GNN-intro}
While GNNs is a broad term encompassing various network structures for graph learning, GCNs \citep{Max} form a specific class of GNNs. GCNs take the response vector $\mathbf{Y}\in\R^{n}$ and the adjacency matrix $A\in\R^{n \times n}$ as the input. Let $L$ be the number of GCN layers. For each $\ell\in\{0,\ldots,L-1\}$, the output of the layers can be defined recursively via
$$H_{\ell+1}=g(H_{\ell},A),$$
where $g$ is a non-linear function and $H_{0}=\mathbf{Y}$. In layer $\ell$, the network has performed a $\ell$-fold composition of the map $g(\cdot,A)$ and we will set $g(\mathbf{Y},A)^{\circ \ell}:=H_\ell.$ A GCN with $L$ layers is then a function $\mathbf{Y}\mapsto g(\mathbf{Y},A)^{\circ L}.$

Specific GCN models differ in the choice and parametrization of the function $g$. An important case is the layer-wise propagation rule with $g$ of the form
\begin{equation}\label{update-rule-function}
\mathbf{y}\mapsto g_\sigma(\mathbf{y},A):=\sigma(T\mathbf{y}W).
\end{equation}
Here, $\sigma$ denotes the component-wise application of the ReLU activation function, $\mathbf{y}\in\R^{n}$ represents a realization of the random vector $\mathbf{Y}$, and $W\in\mathbb{R}$ is a weight parameter. The convolutional operator $T$ is defined as 
\begin{equation}\label{def-T}
T:=\tilde{D}^{-1/2}\tilde{A}\tilde{D}^{-1/2},
\end{equation}
where $\tilde{A}:=A+I_n$ is the adjacency matrix with added self-loops, and $\tilde D$ is the diagonal $n\times n$ matrix with entries $\tilde D_{ii}=\sum_{j}\tilde A_{ij}=d_i+1.$

The above definition can be extended to the case where, at each node, we observe not just one number but a feature vector of length $p$. In this case, $\mathbf{Y}$ becomes an $n\times p$ matrix, and $W$ is a $p\times p$ matrix. For simplicity, we focus on the $p=1$ case for the rest of the paper.

The formulation of \eref{update-rule-function} originates from \cite{Max}. Its motivation is to provide a first-order approximation for the localized spectral filters introduced in \cite{Defferrard} for signal processing, suggesting that GCNs should be applied in the fixed design framework. 

In addition to the widely used propagation operator $T$, this paper also discusses an alternative normalization approach (known as the GraphSAGE convolution \citep{hamilton2017inductive}), that replaces $T$ by
\begin{equation}\label{def-S}
S:=\tilde{D}^{-1}\tilde{A}.
\end{equation}

If the activation function $\sigma$ in \eqref{update-rule-function} is set to the identity function, then the GCN of depth $L$ takes on a specific form
\begin{align}
    g_{\id}(\mathbf{y},A)^{\circ L}
    &= P^L \, \mathbf{y} \, W_{L-1} \ldots W_{0}\label{gcn-id}
\end{align}
with $P=T$, or $P=S$, and $W_0,\ldots,W_{L-1}\in\mathbb{R}$ are the weight parameters in each layer. Empirical studies demonstrate that, compared to nonlinear activation functions \eqref{update-rule-function}, linear GCNs as defined in \eqref{gcn-id} do not have a detrimental effect on accuracy in many downstream applications \citep{pmlr-v97-wu19e}. In \cite{oono2019graph}, the authors provide theoretical evidence that non-linearity does not enhance the expressive power of GCNs. Further discussion comparing linear GCNs to the original GCNs can be found in \cite{gfnn} and \cite{gfnnICPR}.

Rewriting \eqref{gcn-id} yields
\begin{align}
    g_{\id}(\mathbf{y},A)^{\circ L}
    &=P^L \, \mathbf{y} \, W\label{gcn-id-onepara}
\end{align}
with some $W\in\mathbb{R}$. We denote the class of functions described in \eqref{gcn-id-onepara} as $\mathcal{F}_{\text{GCN}}(L)$ and base our analysis later on this class.

\paragraph{The Theory of GCNs.} Several studies have begun investigating the capabilities and limitations of Graph Neural Networks (GNNs), an umbrella term that includes Graph Convolutional Networks (GCNs) as an important example. Most analyses focus on comparing different classes of GNN architectures by using the Weisfeiler-Lemann test to distinguish between graph structures \citep{xu2018powerful} or on performance in classification tasks \citep{garg2020generalization}. While issues such as over-smoothing in deeper GNNs have been explored \citep{li2018deeper, oono2019graph}, there is a noticeable gap in evaluating the consistency of GNN-based estimators. In particular, little has been done to analyze the bias-variance trade-off concerning neighborhood size and topology. This gap highlights the need for a deeper understanding of how graph convolution operations affect the statistical properties of GCNs, motivating our investigation.

\section{THEORETICAL RESULTS FOR DEEP LINEAR GCNS}
Although many variants of GCNs include some form of regularization, the original design incorporates graph information within the layers and deliberately avoids penalization \citep{Max} suggesting that minimizing the least-squares objective over the class of GCNs is sufficient:
\begin{align}
\mathbf{\hat f}_{L}:=\argmin_{g\in\cF_{\text{GCN}}(L)}\big\|\mathbf{Y}-g(\mathbf{Y})\big\|_2^2=\widehat W P^L\mathbf{Y}\label{estimation-procedure}
\end{align}
with $$\widehat W=\frac{\mathbf{Y}^\top P^L\mathbf{Y}}{\|P^L\mathbf{Y}\|_2^2}
=\frac{\mathbf{Y}^\top P^L\mathbf{Y}}{\mathbf{Y}^\top(P^{L})^{\top} P^{L}\mathbf{Y}},$$
where the equality follows by minimizing over the GCN parameter $W$ in \eqref{gcn-id-onepara}. This estimator denoises $Y_i$ by regressing it on itself and neighboring nodes. The size of the neighborhood is controlled by the depth $L.$

To understand the role of the parameter in the GCN, we disregard its randomness and focus on a fixed parameter $W\in\mathbb{R
}.$ For any matrix $P\in\R^{n\times n},$ using that the measurement noise $\eps_i$ in Model \eqref{eq:model} is centered and the triangle inequality, we then obtain the bias-variance type decomposition 
\begin{align}
&\E\cro{\|WP\mathbf{Y}-\mathbf{f^{*}}\|_{2}^2}\nonumber\\
&=\E\cro{\|WP(\mathbf{f^{*}}+{\bm{\eps}})-\mathbf{f^{*}}\|_{2}^{2}}\nonumber\\
&=\|(WP-I_{n})\mathbf{f^{*}}\|_{2}^{2}+W^2\E\cro{\|P{\bm{\eps}}\|_{2}^{2}} \label{decom-equa}\\
&\leq \Big(|W|\cdot\big\|(P-I_{n})\mathbf{f^{*}}\big\|_{2}+|1-W|\cdot \|\mathbf{f^{*}}\|_{2}\Big)^2 \nonumber\\
&\quad  +W^2\E\cro{\|P{\bm{\eps}}\|_{2}^{2}}.\nonumber
\end{align}
Optimizing the parameter $W$ balances the squared bias and variance. In the case of a small signal $\mathbf{f^{*}},$ the optimal $W$ will be in $[0,1)$, thereby inducing shrinkage. An extreme case occurs when $\mathbf{f^{*}}={\bm{0}}.$ Choosing $W=0$ then leads to a vanishing mean squared error. 

In nonparametric statistics, the convergence rate of an estimator depends on the smoothness of the regression function. In the context of graphs, this smoothness is naturally described in terms of the neighborhood structure. As mentioned, we consider graphs $\mathcal{G}$ with the property that neighboring nodes are expected to have similar signals, that is, $f_i^* \approx f_j^*$, whenever $(i,j)\in\mathcal{E}.$ However, the specific similarity assumption concerning neighboring nodes depends on the graph aggregation operator $P$, which incorporates different normalization approaches. We present these formally as follows.
\begin{ass}\label{smoothness-ass}
There exists a $\Delta>0$ such that for any $i$ and any $j\in\cN(i),$
\begin{equation}
\left|f_i^{*}-f_j^{*}\right|\leq \Delta, \quad \text{if} \ P=S,
\label{eq.fiw}
\end{equation}
and
\begin{equation}
\left|\sqrt{d_i+1} \, f_i^{*}-\sqrt{d_j+1} \, f_j^{*}\right|\leq \Delta, \quad \text{if} \ P=T.\label{eq.fiw2}
\end{equation}
\end{ass}
Recall that $\|\cdot\|_{F}$ denotes the Frobenius norm of a matrix. The subsequent result bounds the mean squared error as the sum of the squared bias and variance.
\begin{thm}
\label{thm.main}
If \eqref{eq.fiw} holds, then
\begin{align*}
\frac{1}{n}\E\cro{\|WS^L\mathbf{Y}-\mathbf{f^{*}}\|_{2}^{2}}\leq \Bias_{S}^2(L,W)+\underbrace{W^2\frac{\|S^L\|_{F}^2}{n}}_{\text{variance}},
\end{align*}
and if \eqref{eq.fiw2} holds, then
\begin{align*}
&\frac{1}{n}\E\cro{\|WT^L\mathbf{Y}-\mathbf{f^*}\|_{2}^{2}}
\leq\Bias_{T}^2(L,W)+\underbrace{W^2\frac{\|T^L\|_{F}^2}{n}}_{\text{variance}},
\end{align*}
where $$\Bias_{S}(L,W)=|W|\cdot L\Delta+\frac{|1-W|}{\sqrt{n}}\cdot \|\mathbf{f^{*}}\|_{2},$$ $$\Bias_{T}(L,W)=|W|\cdot L\Delta\sqrt{\frac{2|\mathcal{E}|}{n}+1}+\frac{|1-W|}{\sqrt{n}}\cdot \|\mathbf{f^{*}}\|_{2}.$$
\end{thm}
While the bias terms $\Bias_{S}(L,W)$ and $\Bias_{T}(L,W)$ grow with $L$, Section~\ref{various-exa} demonstrates that the variance terms, specifically $\|S^L\|_{F}^2$ and $\|T^L\|_{F}^2$, can decrease rapidly as $L$ increases. Overall, Theorem~\ref{thm.main} indicates a bias-variance type trade-off mediated by the number of layers $L.$ It implies the existence of an optimal value of $L$ that balances these two terms, providing theoretical evidence for the over-smoothing phenomenon in the regression setting. We will further validate this theoretical finding through an experimental study in Section \ref{sim}, see also Figure~\ref{fig:gtf_latent_L}.

As shown in \eqref{estimation-procedure}, GCNs with linear layers denoise the response variables by averaging over neighboring nodes. A natural competitor is the estimator that estimates the $i$-th signal $f_i^{*}$ by uniformly averaging over the set $\cN^{L}(i),$  which represents the neighboring nodes within a distance of at most $L$ from node $i$. Specifically, the local average estimator of $\mathbf{f^{*}}$ is given by
\begin{align}
     \mathbf{\overline f}_L:=\big(\overline f_{L,1}, \ldots, \overline f_{L,n} \big)^\top,
     \label{eq.loc_mean}
\end{align}
with
\begin{align*}
    \overline f_{L,i}:=\frac{1}{|\cN^{L}(i)|}\sum_{j \in \cN^{L}(i)} Y_j, \quad i=1,\ldots,n.
\end{align*}
The following result bounds the mean squared error of $\mathbf{\overline f}_L$ using a similar bias-variance type decomposition.
\begin{thm}
\label{thm.main2}
For $i=1,\ldots,n,$ the variance of each $\overline{f}_{L,i}$ is given by $\Var\big(\overline{f}_{L,i}\big) =|\cN^{L}(i)|^{-1}$. If \eqref{eq.fiw} holds, then
\begin{equation*}
\frac{1}{n}\E\cro{\|\mathbf{\overline f}_L-\mathbf{f^{*}}\|_{2}^{2}}\leq\underbrace{L^2\Delta^2}_{\text{bias}}+\underbrace{\frac 1n \sum_{i=1}^n\frac 1{|\cN^{L}(i)|}}_{\text{variance}}.
\end{equation*}
\end{thm}
By comparing the bounds in Theorem~\ref{thm.main} and~\ref{thm.main2}, under the same smoothness assumption, the bias term $\Bias_{S}(L,W)$ may be smaller than that of $\mathbf{\overline{f}}_L$, depending on the choice of $W$. We analyze the variance term in details in the next session.

\subsection{Bounds for the Variance via Weighted Local Walks on the Graph}\label{various-exa}
We examine in this section how the network depth $L$ and the local network topology influence the variance decay for the aggregation operators $S$ and $T$.

To this end, we introduce walk analysis as a novel approach. Recall that the definitions of $T$ in \eqref{def-T} and $S$ in \eqref{def-S} rely on the adjacency matrix with added self-loops, $\tilde A$. While the original graph is denoted by $\mathcal{G}$, we write $\mathcal{G}'$ for the graph augmented with self-loops. A path from node $i$ to node $j$ via nodes  $\ell_1,\ldots,\ell_{L-1} \in \{1,\ldots,n\}$ is denoted by $$(i\to\ell_1\to\ldots\to\ell_{L-1}\to j).$$ 
The length of the path is $L.$ The weights associated with this path under the aggregation operator $S$ are defined as 
$$\omega_{i}(\ell_1,\ldots,\ell_{L-1}):=\frac{1}{(d_i+1)(d_{\ell_1}+1)\cdot \ldots \cdot (d_{\ell_{L-1}}+1)}$$ 
and under $T$ as
\begin{align*}
\widetilde \omega_{ij}(\ell_1,\ldots,\ell_{L-1}):=\Big(\frac{d_i+1}{d_j+1}\Big)^{1/2}\omega_{i}(\ell_1,\ldots,\ell_{L-1}),
\end{align*}
where $d_i$ is the edge degree of node $i$. Let $\mathcal{P}_L(i \to j)$ denote all paths from node $i$ to node $j$ of length $L$ in the self-loop augmented graph $\mathcal{G}'.$ Each path of length $L$ in the augmented graph corresponds to a path of length $\leq L$ in the original graph $\mathcal{G}.$ The edge degrees and the length of the paths $L$ decrease the weights but increase the number of paths. The key observation is that for any vector $\mathbf{v}=(v_1,\ldots,v_n)^\top,$
\begin{align*}
\big(S^L \mathbf{v}\big)_i=\sum_{j=1}^{n}\sum_{(i\to\ell_1\to\ldots\to\ell_{L-1}\to j)\in \mathcal{P}_L(i \to j)}\hspace{-10mm}v_j\omega_{i}\big(\ell_1,\ldots,\ell_{L-1}\big)
\end{align*}
and 
\begin{align*}
\big(T^L \mathbf{v}\big)_i=\sum_{j=1}^{n}\sum_{(i\to\ell_1\to\ldots\to\ell_{L-1}\to j)\in \mathcal{P}_L(i \to j)}\hspace{-10mm}v_j\widetilde \omega_{ij}\big(\ell_1,\ldots,\ell_{L-1}\big),
\end{align*}
with a detailed derivation provided in the appendix.

Since the noise vector $\bm{\eps}=(\eps_1,\ldots, \eps_n)^\top$ consists of centered and uncorrelated random variables with variance one, we find
\begin{align*}
\E\big[\big(S^L \bm{\eps}\big)_i^2\big]=\sum_{j=1}^{n}\cro{\sum_{(i\to\ell_1\to\ldots\to\ell_{L-1}\to j)\in \mathcal{P}_L(i\to j)}\hspace{-17mm}\omega_{i}(\ell_1,\ldots,\ell_{L-1})}^2
\end{align*}
and 
\begin{align*}
\E\big[\big(T^L \bm{\eps}\big)_i^2\big]=\sum_{j=1}^{n}\cro{\sum_{(i\to\ell_1\to\ldots\to\ell_{L-1}\to j)\in \mathcal{P}_L(i\to j)}\hspace{-17mm}\widetilde \omega_{ij}(\ell_1,\ldots,\ell_{L-1})}^2.
\end{align*}
If the components of $\bs{\eps}$ are uncorrelated and have variance one, then for $P \in \{S, T\},$ we have $$\|P^L\|_{F}^2=\E\cro{\|P^{L}{\bs{\eps}}\|_{2}^{2}}=\sum_{i=1}^{n}\E\cro{\big(P^L \bm{\eps}\big)_i^2}.$$
The quantity $\E[(P^L \bm{\eps})_i^2]$ thus represents the contribution of node $i$ to the variance term in Theorem~\ref{thm.main} after $L$ applications of the propagation operator $P$.

We first compare $\E[(S^L \bm{\eps})_i^2]$ with the variance $\Var(\overline{f}_{L,i})$ of the uniform average estimator $\overline{f}_{L,i}$ defined in \eqref{eq.loc_mean}. Since the row sums of the matrix $S=\tilde D^{-1}\tilde A$ are all one, $S$ is a transition matrix of a Markov chain. Thus, $S^L$ is a transition matrix with row sums equal to one and for any node $i\in\{1,\ldots,n\},$
$$\sum_{j=1}^{n} \ \sum_{(i\to\ell_1\to\ldots\to\ell_{L-1}\to j) \in \mathcal{P}_L(i \to j)}\omega_{i}(\ell_1,\ldots,\ell_{L-1})=1.$$
In fact, in the previous display, $\sum_{j=1}^{n}$ can be replaced by $\sum_{j\in \cN^{L}(i)}.$ Applying the Cauchy-Schwarz inequality (AM-QM) and Theorem \ref{thm.main2}, we deduce 
\begin{align}
    \E\big[\big(S^L \bm{\eps}\big)_i^2\big]&\geq\frac{1}{|\cN^{L}(i)|}=\Var\big(\overline{f}_{L,i}\big) \, .
    \label{lower-bound-var}
\end{align}
For $L=1,$ equality holds in \eqref{lower-bound-var}. The inequality shows that, in general, the local averaging estimator $\mathbf{\overline f}_L$ has smaller variance if we take the same $L$ in the GCN estimator and the local averaging estimator. On the contrary, the GCN estimator is expected to have a smaller bias, as it computes a weighted sum over the local neighborhood $\cN^{L}(i),$ assigning more mass to nodes that are closer to $i$ in path distance. Although the gain in the bias is hard to quantify theoretically, we see that it outweighs the loss in the variance in our empirical studies. For instance, in all the scenarios in Figure \ref{fig:regression_results} the GCN estimator has smaller test error.

In the remainder of this section, we study the decay rate of $\E[(S^L \bm{\eps})_i^2]$ with respect to the depth $L$. Notably, the decay can vary significantly depending on the local graph structure around node $i$. Since $S$ and $T$ behave similarly, we focus on the results for $\E[(S^L \bm{\eps})_i^2]$ and delay the analysis of $\E[(T^L \bm{\eps})_i^2]$ to the appendix.

We first provide an upper bound assuming that node $i$ is the root of a locally-rooted tree. In this scenario, the variance term decays exponentially in the number of layers $L$.

\begin{prop}\label{tree-structure}
Assume the local graph around node $i$ is a rooted tree, in the sense that the subgraph induced by the vertices $\cN^L(i)$ forms a  rooted tree with root $i$ and all nodes having edge degree $d$. Then, we have 
\begin{align*}
    \E\big[\big(S^L \bm{\eps}\big)_i^2\big]\leq 4(d+1)^{-L} (L+1)3^{2L}.
\end{align*}
\end{prop}
In this case, the variance decay in $d$ is $(d+1)^{-L}.$ This is the sharp rate, since the lower bound in \eqref{lower-bound-var} gives the same rate, namely $1/|\cN^L(i)| = (d-1)/(d^{L+1}-1)\approx d^{-L}$ for large $d$.

A slow rate of decay for the variance occurs if node $i$ has edge degree $d$ but is connected to nodes with small edge degrees.
\begin{prop}\label{tree-structure-slow}
If there exists a path from node $i$ with degree $d$ to node $j$ via nodes $\ell_1,\ldots,\ell_{L-1}\in\{1,\ldots,n\}$ such that $d_{\ell_1}, \ldots,  d_{\ell_{L-1}}\leq 3,$ then \begin{align*}
    \E\big[\big(S^L \bm{\eps}\big)_i^2\big]
    \geq (d+1)^{-2}4^{2-2L}.
\end{align*}
\end{prop}
This shows that for large $d$ and large $L$, the variance decay is necessarily much slower than the rate $(d+1)^{-L}$ obtained in Proposition~\ref{tree-structure}.

The next result shows that adding a cycle to node $i$ can immediately slow variance decay.
\begin{prop}\label{circle-lem}
Assume that the graph decomposes into a cycle $\mathcal{C}$ of length $r$ and a graph $\mathcal{H}$ in the sense that $\mathcal{C}$ and $\mathcal{H}$ share exactly one node $i:=\mathcal{C}\cap\mathcal{H}$ and there are no edges connecting $\mathcal{C}\setminus\{i\}$ and $\mathcal{H}\setminus\{i\}.$ Then, for any $L=1,2,\ldots$,
\begin{equation}\label{circle-slow-down}
\E\big[\big(S^L \bm{\eps}\big)_i^2\big]\geq \cro{\frac{3}{(d_i+1)(r-1)}}^21.5^{-2L}.
\end{equation}
\end{prop}

The variance term in GCNs is sensitive to small perturbations in the graph. To see this, suppose node $i$ admits a local tree structure as described in Proposition~\ref{tree-structure}. Attaching an additional cycle to node $i$, \eqref{circle-slow-down} demonstrates that regardless of the edge degree $d_i$ of node $i$, the variance decay rate in $(d_i,L)$ will immediately drop from $\lesssim  (d_i+1)^{-L}$ to $\gtrsim (d_i+1)^{-2} 2.25^{-L}$. On the contrary, the local average estimator \eqref{eq.loc_mean} will hardly be affected by an additional cycle. 

Proposition~\ref{circle-lem} serves as a prototype illustrating why stacking many GCN layers often results in poor performance. In practical applications like social network analysis, nodes frequently exhibit additional cycles. Proposition~\ref{circle-lem} suggests that in such cases, more GCN layers are needed to balance the bias-variance trade-off. However, as shown in Theorem~\ref{thm.main}, increasing the number of layers also raises the variance term, ultimately leading to the well-known over-smoothing phenomenon \citep{oono2019graph,cai2020,rusch2023,li2018deeper}. This phenomenon contrasts sharply with Fully Connected Neural Networks (FNNs), where both theory and empirical evidence suggest that greater depth increases the network expressiveness \citep{Matus,10.1214/19-AOS1875}.

\section{NUMERICAL SIMULATIONS}\label{sim}
We validate our results through a series of numerical experiments\footnote{The code used for the experiments in the following two sections is available at the following link: \url{https://github.com/donnate/GNN_regression}. }. Specifically, to evaluate the link between neighborhood topology -- particularly degree distribution and degree skewness -- and optimal neighborhood size, we generate graphs with 
$n=100$ nodes using the following topologies:
\begin{itemize}[noitemsep]
    \item \textit{Latent variable graphs}: For each node $i$,  we generate a latent variable vector $U_{i\cdot}\in\mathbb{R}^{1\times2}$ by sampling from a two-dimensional uniform distribution on $[0,1]^2$. The edges of the network are then sampled from a Bernoulli distribution with $p_{ij} = 1/( 1 + e^{-5  \| U^{\top}_{i\cdot} - U^{\top}_{j\cdot}\|_2})$, where the scaling factor (here, $5$) is chosen empirically to control the edge density. To further measure the effect of the edge density on the optimal neighborhood size, we sparsify the induced network as follows. We first sample a Minimum Spanning Tree (MST) as the graph's ``backbone'' to ensure the sparsified graph remains connected. We then delete edges that are not in the MST with probability $p$, so that a higher $p$ results in a sparser graph. 
    \item \textit{Preferential attachment graphs:} the graphs are generated using the Holme and Kim algorithm \citep{holme2002growing}, varying their local clustering probability $p$. In this case, higher values of $p$ yield graphs with higher clustering coefficients.
    \item \textit{k-regular trees:} where $k$ denotes the degree of each node in the tree.
   \item \textit{Barbell graphs:} consisting of two cliques of $m$ fully connected nodes, joined by a chain of $n - 2m $ nodes.
\end{itemize} 
For the last three graph topologies, we create a latent variable $U_{i\cdot}\in\mathbb{R}^{1\times2}$ for each node by taking the coordinates of the first two eigenvectors of the graph Laplacian. Denoting the eigenvalue decomposition of the graph Laplacian ${\bf L} = D - A$ as ${\bf L}= {\bf V} \Lambda {\bf V}^{\top}$, we define $\bf U$ as the matrix formed by the first two columns of $\bf V$, i.e., ${\bf U} = {\bf V}[,1:2].$

A node signal $\bf f^{*}$ is then generated as a smooth function of the latent variable vector $U_{i\cdot}\in \mathbb{R}^{1\times2}$: \begin{equation}\label{exp-sig-design}
f_{i}^* = 2\cos(U_{i\cdot}\beta),   \end{equation} where $U_{i\cdot}$ denotes the $i$-th row of ${\bf U}$ and $\beta$ is a vector of coefficients of the form $\beta = (-\alpha,\alpha)^{\top}$, with $\alpha$ adjusted to control different levels of smoothness across the graph. As highlighted in Figures \ref{fig:enter-label-2}-\ref{fig:enter-label-5}, lower values of $\alpha$ tend to create smoother functions, while higher values of $\alpha$ induce faster varying signals. 

The left subplots of Figures \ref{fig:enter-label-2}-\ref{fig:enter-label-5} present examples of the generated graphs, with node colors indicating the true $\mathbf{f}^*$. The middle and right subplots provide examples of the graph's spectral embeddings (or latent variables) ${\bf U}$, for different values of $\alpha$, showcasing signals with various levels of smoothness. Finally, the observed response vector $\mathbf{Y}=(Y_1,\ldots,Y_n)$ is simply generated as:
$$Y_i=f_i^* +\varepsilon_i$$
where $\varepsilon_i \sim N(0, \sigma^2)$ is independent Gaussian noise. Unless otherwise specified, $\sigma$ is set to 1. In all subsequent experiments, consistent with the GNN literature on transductive learning (e.g., \cite{Max}), we partition the dataset, using 20\% of the nodes for training and a separate 20\% for testing. This emulates the standard GNN setting, where {\it only a fraction of the nodes} is available for training. We compare the results of the estimators obtained by (a) smoothing the signal on neighborhoods of size $L$ (where $L$ varies), using either the GCN ($T$) or GraphSAGE ($S$). These estimators have no learned parameters and solely assess the effect of neighborhood smoothing, which we compare to the local averaging estimator $\mathbf{\overline f}_L$; and (b) trained GNNs (either the GCN or GraphSAGE) with a hidden dimension of size 8 and non-linearities. The purpose of this second set of methods is to show that our results extends to trained, non-linear GNNs.

\begin{figure*}[t]
    \centering
\includegraphics[width=\linewidth,height=3.31cm]{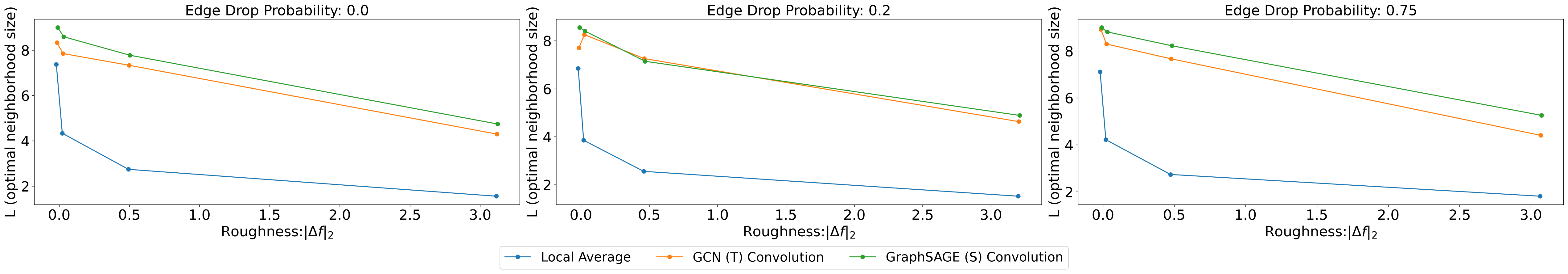}
\caption{Optimal number of convolutions as a function of the roughness $\|\Delta {\bf f^{*}}\|_2 = \sqrt{[\sum_{(i,j) \in \mathcal{E}} (f_i^{*} - f_j^{*})^2]/|\mathcal{E}|}$ of $\mathbf{f}^{*}$ on the latent variable graph with $\sigma^2=2$. Each subplot corresponds to a different sparsification level, with 0 representing the original graph and 0.75 representing a graph with 75\% of its edges removed.}
\label{fig:gtf_latent_L}
\end{figure*}

\paragraph{Result 1: The optimal convolution size $L$ decreases sharply as the signal's roughness increases.} Figure~\ref{fig:gtf_latent_L} displays the optimal neighborhood size, averaged over 20 experiments, as a function of the graph's roughness (simply defined as $\sqrt{[\sum_{ (i,j) \in \mathcal{E}} (f_i^{*} - f_j^{*})^2]/|\mathcal{E}|})$ for the latent variable graph. We observe that the optimal number of convolutions decreases as a function of the graph roughness, as long as the neighborhood is informative of the underlying signal. Similar figures for the other three graph topologies (and for various levels of noise) are presented in Figures~\ref{fig:opt_L_barbell}-\ref{fig:opt_L_PL} in the appendix. The result confirms the theoretical findings presented in Theorem~\ref{thm.main}, which illustrate the bias-variance trade-off phenomenon. Interestingly, the GraphSAGE convolution 
$S$ results in optimal neighborhood sizes that tend to be slightly larger than those of the usual GCN convolution, particularly for very smooth signals. 

\paragraph{Result 2: The variance appears to be generally lower for the GCN convolution ($T$) compared to GraphSAGE ($S$).} Figures \ref{fig:barbell-20}-\ref{fig:latent-0.5} in the appendix illustrate the bias-variance trade-off across different graph topologies as the number of convolutions increases. As expected, and as predicted by \eref{lower-bound-var}, the local averaging estimator consistently exhibits lower variance than the GNN convolutions ($S$ and $T$). Conversely, the bias of the local averaging operator increases more substantially as a function of $L$ than that of the GNN convolutions, particularly in cases where the signal is not extremely smooth. In the Barbell graph, for instance, 
local averaging shows three times the bias of GNNs for $L=4$ at $\alpha=5$. GNNs with learned weights and non-linearities behave similarly for sufficiently smooth signals. Across topologies, the signal ${f_i^*} = 2\cos(U_{i\cdot}\beta)$ with $\beta=(-0.1, 0.1)^{\top}$ yields smooth graphs that display a clear bias-variance trade-off. Moreover, non-linear GCNs tend to exhibit lower variance than their GraphSAGE counterparts.

\begin{figure*}[!ht]
        \centering
\includegraphics[width=\linewidth,height=3.83cm]{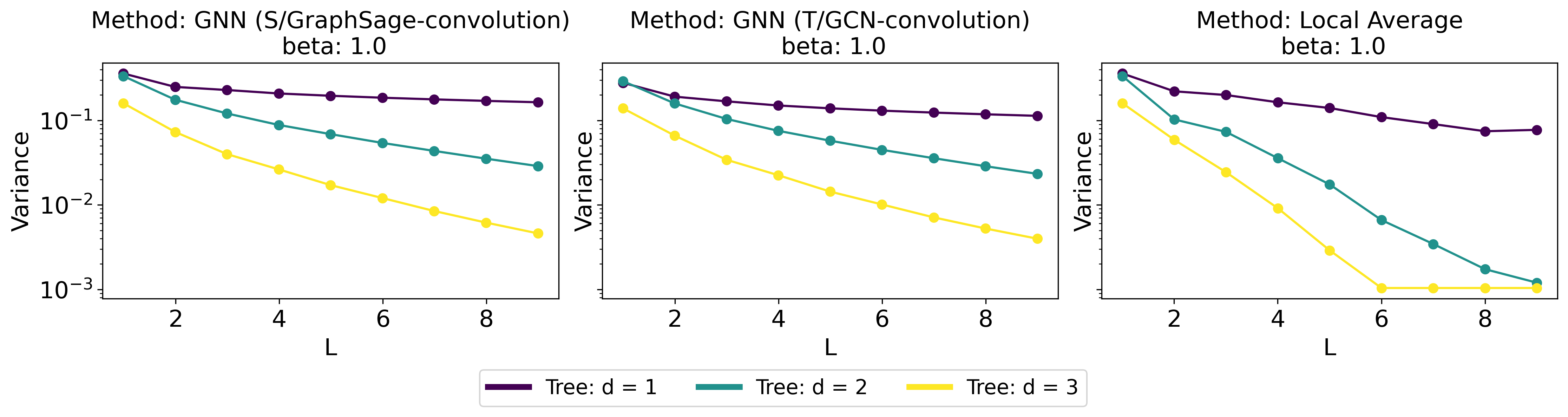}
        \caption{The variance decay at the root of the tree as a function of $L$ under the signal $f_i^* = 2 \cos(U_{i\cdot} \beta)$, with $\beta = (-1,1)^{\top}$, colored by degree value. The $y$-axis is shared across all 3 plots.}
    \label{fig:tree_low}
\end{figure*}

\paragraph{Result 3: The graph topology does affect the rate of variance decay.} The derived theory (Proposition~\ref{tree-structure}) predicts exponential variance decay in rooted trees with respect to the number of layers $L$. Moreover, the larger the degree $d$ of the root node, the faster the variance decay. Figure~\ref{fig:tree_low} shows the decay of the variance as a function of $L$ (logged on the $y$-axis), for the signal $f_i^*$ defined in \eqref{exp-sig-design}. As expected, the log of the variance decays linearly with the number of convolutions $L$, with higher degrees resulting in steeper decay slopes.

Our theory also predicts that, for instance, nodes with a high degree but connected to low-degree nodes will exhibit a lower variance decay than nodes connected to high-degree nodes. To this end, we compare the same signal in settings where either a complete graph or a star is added to a tree. In both settings, the local average estimator is expected to behave similarly. We find (Figure~\ref{fig:star}) that the variance for the GCN convolution on the star graph does not decay effectively.

\begin{figure*}[!ht]
\includegraphics[width=\linewidth,height=3.15cm]{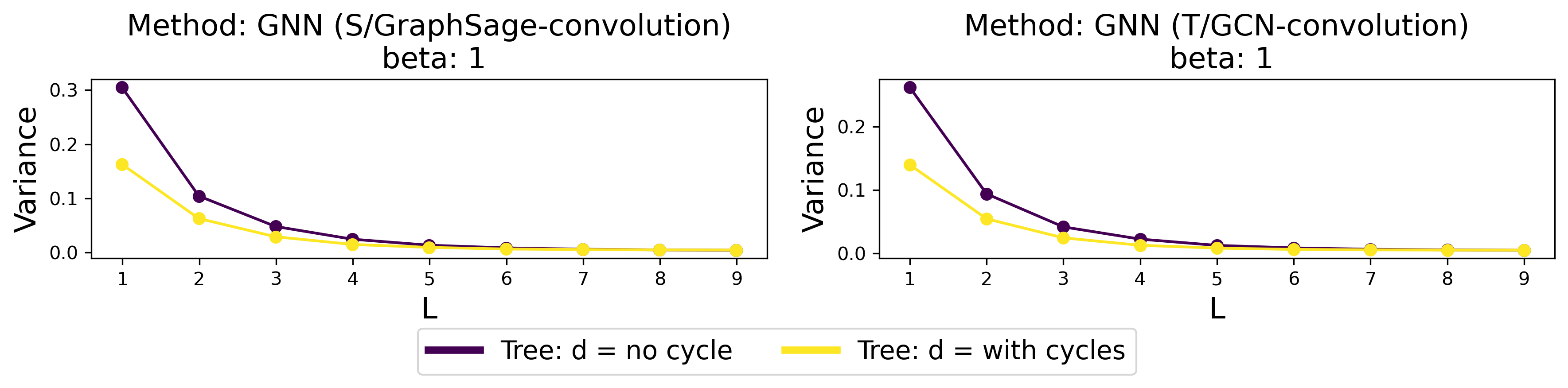}
\caption{The variance decay at the root of the tree as a function of $L$ under the signal $f_i^* = 2 \cos(U_{i\cdot} \beta)$, with $\beta = (-1,1)^{\top}$, and its behavior after adding cycles.}
\label{fig:tree_cycles}
\end{figure*}

Next, we investigate the impact of adding local cycles, as discussed in Proposition~\ref{circle-lem}. We construct a binary tree and introduce edges between nodes at levels 2 and 3. The results in Figure~\ref{fig:tree_cycles} confirm that variance decays much more slowly with cycles.

\section{REAL-DATA ANALYSIS}
We validate our framework on 6 real-world examples. More specifically, we used the publicly available regression datasets used as validation by \cite{jia2020residual} and \cite{huang2024uncertainty}. These datasets include 3 spatial datasets on election results, income as well as education levels across the counties in the United States. We also include 2 transportation networks, based on traffic data in the cities of Anaheim and Chicago. Each vertex represents a directed lane, and two lanes that meet at the same intersection are connected. In these datasets, the response $Y_i$ represents the traffic flow in the lane. Finally, the Twitch dataset captures
friendships among Twitch streamers in Portugal, while the node signal corresponds to the logarithm of the number of viewers for each streamer. Figure \ref{fig:enter-label} provides details on the datasets and their properties, emphasizing their distinct topological characteristics. In particular, spatial datasets typically have a regular degree distribution, whereas the Twitch dataset exhibits a highly skewed degree distribution.

\begin{figure*}[!h]
\includegraphics[width=\textwidth,height=6cm]{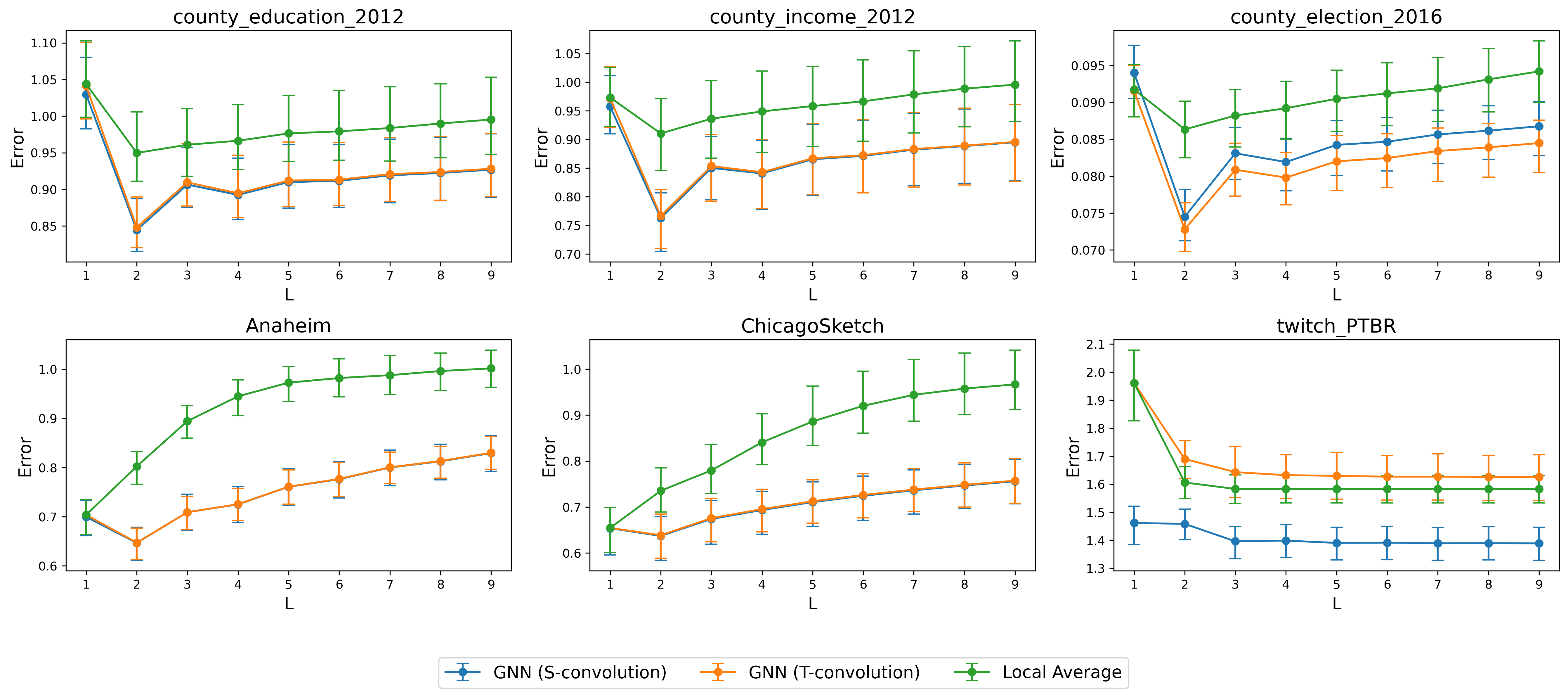}
    \caption{Mean Squared Error (MSE) as a function of the neighborhood size. The nodes denote the mean MSE over 50 random splits of the data into training and validation sets, and the error bars denote interquartile ranges.}\label{fig:regression_results}
\end{figure*}

We use our estimator in the prediction setting, where the value at the node of interest is unobserved. In this setting, we predict $\hat{Y}_i$ using different convolutions: $\hat{Y}_i = \sum_{j\in\mathcal{N}(i)} w_{ij} Y_j$, where the weights in the sum depend on the convolution.  To emulate a real-life training scenario, we randomly select a set of 500 nodes for training and another 500 for validation. Figure~\ref{fig:regression_results} shows the mean squared error on the validation set as a function of the neighborhood size. Interestingly, we observe that the optimal neighborhood size is reached at $L=2$ for all spatial graphs. In these cases, the degree distribution is relatively homogeneous across nodes (see statistics in the appendix), and the convolutions $S$ and $T$ produce fairly similar results. However, in the social network example, different aggregation operators lead to drastically different effects, with the GraphSAGE convolution outperforming the other two.

We also test our estimator as a denoiser. In this setting, we denoise the observed value $Y_i$ as $\hat{Y}_i =\sum_{j\in\mathcal{N}^1(i)} w_{ij} Y_j$, with the weights depending on the convolution. To avoid data snooping between the training and test sets, we replace the values of the test nodes with those of one of their direct neighbors. Assuming that direct neighbors have roughly the same distribution, this modification ensures that no information from the training set permeates the test set while creating another version of the same graph. The results are presented in the appendix in Figure ~\ref{fig:denoising_real}. 

\section{CONCLUSION}
In this work, we lay the groundwork for a statistical understanding of GCNs. GCNs perform weighted local averaging over neighboring nodes, with the neighborhood size determined by the depth $L$ of the network. In the context of denoising, we characterize a bias-variance type trade-off risk bound in terms of $L$ for both the original GCN and GraphSAGE convolution. We showed that the variance term of the GraphSAGE convolution is lower-bounded by its counterpart from the local averaging estimator. We propose investigating the variance term from a new perspective: walk analysis, which enables the derivation of both lower and upper bounds of the variance based on local graph topologies. Our analysis reveals a lack of robustness in GCNs, in that small perturbations in the graph structure can lead to substantial changes in the variance.

\section*{Acknowledgments}
We thank the anonymous reviewers for their helpful comments. The research of J. S.-H. and J. C. was supported by the NWO Vidi grant VI.Vidi.192.021. The research of C. D. was funded by the National Science Foundation (Award Number 2238616), as well as the resources provided by the University of Chicago’s Research Computing Center. The work of O. K. was funded by CY Initiative (grant “Investissements d’Avenir” ANR-16-IDEX0008) and Labex MME-DII(ANR11-LBX-0023-01). Part of this work was carried out while J. S.-H. and J. C. were visiting the Simons Institute for the Theory of Computing in Berkeley.

\renewcommand{\bibsection}{}
\section*{References}
\bibliographystyle{apalike}
\bibliography{bib}

\appendix
\setcounter{equation}{0}
\renewcommand{\theequation}{A.\arabic{equation}}
\newtheorem{innercustomgeneric}{\customgenericname}
\providecommand{\customgenericname}{}
\newcommand{\newcustomtheorem}[2]{%
  \newenvironment{#1}[1]
  {%
   \renewcommand\customgenericname{#2}%
   \renewcommand\theinnercustomgeneric{##1}%
   \innercustomgeneric
  }
  {\endinnercustomgeneric}
}
\newcustomtheorem{customthm}{Proposition}
\onecolumn
\section{PROOFS OF MAIN THEOREMS}
\subsection{Proof of Theorem \ref{thm.main}}
Recall that as the noise $\eps_i$ is centered, with inequality \eref{decom-equa} in the paper, we have for any fixed parameter $W\in\R$ and matrix $P\in\R^{n\times n}$,
\begin{align}
\E\cro{\|WP\mathbf{Y}-\mathbf{f^{*}}\|_{2}^{2}}\leq \Big( |W|\cdot\big\|P\mathbf{f^{*}}-\mathbf{f^{*}}\big\|_{2}+|1-W|\cdot \|\mathbf{f^{*}}\|_{2}\Big)^2+W^2\E\cro{\|P{\bm{\eps}}\|_{2}^{2}}.\label{P-inq}
\end{align}
We first consider the case $P=S^L$ for $L=1,2,\ldots$. To further bound \eqref{P-inq} for $P=S^{L}$, we now prove by induction on $L$ that
\begin{equation}\label{induc-bias}
\|S^{L}\mathbf{f^{*}}-\mathbf{f^{*}}\|_{\infty}\leq L\Delta,
\end{equation}
where $\|{\bf{v}}\|_{\infty}:=\max_{i=1,\ldots,n}|v_i|$ for any vector ${\bf{v}}=(v_1,\ldots,v_n)^{\top}$.

Write
\begin{equation}\label{tilde-A}
    \tilde A=
\begin{pmatrix}
    a_{11}&a_{12}&\ldots&a_{1n}\\
   a_{21}&a_{22}&\ldots&a_{2n}\\
    \ldots&\ldots&\ldots&\ldots\\
    a_{n1}&a_{n2}&\ldots&a_{nn}\\
\end{pmatrix},
\end{equation}
where $a_{ii}=1$, for all $i\in\{1,\ldots,n\}$ and $a_{ij}=a_{ji}\in\{0,1\}$, for $i\not=j,$ indicating whether nodes $i$ and $j$ are connected by an edge belonging to $\mathcal{E}$. The matrix $\tilde A$ represents the connection structure of the augmented graph $\mathcal{G}'$, where self-loops are included by setting $a_{ii}=1$, for $i=1,\ldots,n$. Thus,
\begin{equation}\label{S-for}
S=\tilde{D}^{-1}\tilde{A}=
\begin{pmatrix}
    \frac{a_{11}}{d_{1}+1}&\frac{a_{12}}{d_{1}+1}&\ldots&\frac{a_{1n}}{d_{1}+1}\\
   \frac{a_{21}}{d_{2}+1}&\frac{a_{22}}{d_{2}+1}&\ldots&\frac{a_{2n}}{d_{2}+1}\\
    \ldots&\ldots&\ldots&\ldots\\
    \frac{a_{n1}}{d_{n}+1}&\frac{a_{n2}}{d_{n}+1}&\ldots&\frac{a_{nn}}{d_{n}+1}\\
\end{pmatrix},
\end{equation}
where $d_i$ is the edge degree of node $i$.

For $L=1$, we can employ the smoothness condition \eref{eq.fiw} in Assumption~\ref{smoothness-ass} and the fact that the edge degree $d_i$ is the same as the number of neighbors of $i,$ to show
\begin{align*}
\|S\mathbf{f^{*}}-\mathbf{f^{*}}\|_{\infty}&=\|\tilde D^{-1}\tilde A \mathbf{f^{*}}-\mathbf{f^{*}}\|_{\infty}\\
&=\max_{i=1,\ldots,n}\left|\sum_{j=1}^{n}\frac{a_{ij}}{d_{i}+1}f_j^{*}-f_i^{*}\right|\\
&=\max_{i=1,\ldots,n}\left|\sum_{j\in\cN(i)}\frac{1}{d_{i}+1}\big(f_j^{*}-f_i^{*}\big)\right|\\
&\leq\Delta.
\end{align*}
For the induction step, assume the claim holds until $L-1$, in the sense that, for any $1\leq \ell\leq L-1$, $\|S^{\ell}\mathbf{f^{*}}-\mathbf{f^{*}}\|_{\infty}\leq \ell \Delta.$ Then,
\begin{align*}
    \|S^{L}\mathbf{f^{*}}-\mathbf{f^{*}}\|_{\infty}&=\big\|S(S^{L-1}\mathbf{f^{*}}-\mathbf{f^{*}})+S\mathbf{f^{*}}-\mathbf{f^{*}}\big\|_{\infty}\\
    &\leq\big\|S(S^{L-1}\mathbf{f^{*}}-\mathbf{f^{*}})\big\|_{\infty}+\|S\mathbf{f^{*}}-\mathbf{f^{*}}\|_{\infty}\\
&\leq\max_{i=1,\ldots,n}\cro{\sum_{j=1}^{n}\frac{a_{ij}}{d_{i}+1}(L-1)\Delta}+\Delta\\
&=(L-1)\Delta\max_{i=1,\ldots,n}\cro{\sum_{j\in\cN(i)}\frac{1}{d_{i}+1}}+\Delta\\
&\leq L\Delta,
\end{align*}
completing the induction step. Thus, we conclude that $\|S^{L}\mathbf{f^{*}}-\mathbf{f^{*}}\|_{\infty}\leq L\Delta,$ for all $L=1,2,\ldots$. This implies that 
\begin{equation}\label{bias-S}
\|S^L\mathbf{f^{*}}-\mathbf{f^{*}}\|_{2}\leq\sqrt{n}\|S^L\mathbf{f^{*}}-\mathbf{f^{*}}\|_{\infty}\leq \sqrt{n}L\Delta.
\end{equation}
Moreover, for the variance part, we use the fact that the components of $\bs{\eps}$ are uncorrelated and have a variance of one to obtain 
\begin{equation}\label{var-S}
\E\cro{\|S^{L}{\bs{\eps}}\|_{2}^{2}}=\Tr((S^L)^{\top}S^L)=\|S^L\|_{F}^2.
\end{equation}  
Plugging \eqref{bias-S} and \eqref{var-S} into \eqref{P-inq} with $P=S^L$, we deduce that
\begin{align*}
    \frac{1}{n}\E\cro{\|WS^{L}\mathbf{Y}-\mathbf{f^{*}}\|_{2}^{2}}&\leq\frac{1}{n}\Big(|W|\cdot\big\|S^{L}\mathbf{f^{*}}-\mathbf{f^{*}}\big\|_{2}+|1-W|\cdot \|\mathbf{f^{*}}\|_{2}\Big)^2+\frac{W^2}{n}\E\cro{\|S^{L}{\bm{\eps}}\|_{2}^{2}}\\
    &\leq\left(|W|\cdot L\Delta+\frac{|1-W|}{\sqrt{n}}\cdot \|\mathbf{f^{*}}\|_{2}\right)^2+W^2\frac{\|S^L\|_{F}^2}{n}\\
    &=\Bias_{S}^2(L,W)+W^2\frac{\|S^L\|_{F}^2}{n},
\end{align*}
which concludes the proof of the first upper bound in Theorem~\ref{thm.main} of the paper.

Next, we address the case $P=T^{L}$ with $T=\tilde D^{-1/2}\tilde{A}\tilde D^{-1/2}.$ Observe that for $L=1,2,\ldots,$
$$T^L=\tilde D^{1/2} S^L\tilde D^{-1/2},$$
which implies
\begin{align}
    T^L \mathbf{f^{*}}-\mathbf{f^{*}}
    &=\tilde D^{1/2}S^L\tilde D^{-1/2} \mathbf{f^{*}}-\mathbf{f^{*}}
    =\tilde D^{1/2}\big(S^L -I_{n}\big) \tilde D^{-1/2}\mathbf{f^{*}}.\label{SL-TL-trans}
\end{align}
Since $\mathbf{f^{*}}$ satisfies \eref{eq.fiw2} in Assumption~\ref{smoothness-ass}, the vector $\tilde D^{-1/2} f^{*}$ satisfies \eref{eq.fiw} in Assumption~\ref{smoothness-ass}. Using \eqref{induc-bias} and \eqref{SL-TL-trans},
\begin{align}
\|T^L \mathbf{f^{*}}-\mathbf{f^{*}}\|_2&=\|\tilde D^{1/2}\big(S^L -I_{n}\big) \tilde D^{-1/2}\mathbf{f^{*}}\|_{2}\nonumber\\
&\leq\sqrt{\sum_{i=1}^n (d_i+1)}\cdot\|\big(S^L -I_n\big) \tilde D^{-1/2} \mathbf{f^{*}}\|_\infty\nonumber\\
&\leq\sqrt{\sum_{i=1}^n (d_i+1)}\cdot L\Delta\nonumber\\
&=\sqrt{2|\mathcal{E}|+n}\cdot L\Delta,\label{bias-T}
\end{align}
where the last equality follows from $|\mathcal{E}|=(\sum_{i=1}^n d_i)/2$.

For the variance term, we can proceed similarly as in \eqref{var-S} and obtain
\begin{equation}\label{var-T}
\E\cro{\|T^{L}{\bs{\eps}}\|_{2}^{2}}=\Tr((T^L)^{\top}T^L)=\|T^L\|_{F}^2.
\end{equation}
Plugging \eqref{bias-T} and \eqref{var-T} into \eqref{P-inq} with $P=T^L$, we deduce
\begin{align*}
  \frac{1}{n}\E\cro{\|WT^{L}\mathbf{Y}-\mathbf{f^{*}}\|_{2}^{2}}&\leq\frac{1}{n}\Big(|W|\cdot\big\|T^{L}\mathbf{f^{*}}-\mathbf{f^{*}}\big\|_{2}+|1-W|\cdot \|\mathbf{f^{*}}\|_{2}\Big)^2+\frac{W^2}{n}\E\cro{\|T^{L}{\bm{\eps}}\|_{2}^{2}}\\
&\leq\left(|W|\cdot L\Delta\sqrt{\frac{2|\mathcal{E}|}{n}+1}+\frac{|1-W|}{\sqrt{n}}\cdot \|\mathbf{f^{*}}\|_{2}\right)^2+W^2\frac{\|T^L\|_{F}^2}{n}\\
&=\Bias_{T}^2(L,W)+W^2\frac{\|T^L\|_{F}^2}{n},
\end{align*}
proving the second upper bound in Theorem~\ref{thm.main}.

\subsection{Proof of Theorem \ref{thm.main2}}

By definition,
$$\overline f_{L,i}=\frac{1}{|\cN^{L}(i)|}\left(\sum_{j \in \cN^{L}(i)} Y_j\right)=\frac{1}{|\cN^{L}(i)|}\cro{\sum_{j \in \cN^{L}(i)}(f_j^*+\eps_{j})}.$$
Since the components of $\bs{\eps}$ are uncorrelated and have a variance of one, we obtain
$$\Var\big(\overline{f}_{L,i}\big)=\frac{1}{|\cN^{L}(i)|^2}\sum_{j \in \cN^{L}(i)}\Var\big(f_j^{*}+\eps_{j}\big)=\frac{1}{|\cN^{L}(i)|}.$$
For any $j\in\cN^{L}(i)$, there exists a path in the augmented graph $\mathcal{G}'$, where self-loops are allowed, given by
\begin{equation}\label{path-L}
(i\to\ell_{1}\to\ldots\to\ell_{L-1}\to j),
\end{equation}
which connects node $i$ to node $j$ through nodes  $\ell_1,\ldots,\ell_{L-1} \in \mathcal{V}$ and has a length of $L$. The path in \eqref{path-L} also implies that
$\ell_{1}\in\cN^{1}(i),\ell_{2}\in\cN^{1}(\ell_{1}),\ldots,j\in\cN^{1}(\ell_{L-1})$. Therefore, if condition \eref{eq.fiw} in Assumption \ref{smoothness-ass} holds, we have, for any $j\in\cN^{L}(i)$,
\begin{align}
    |f_i^*-f_j^*|\leq|f_i^*-f_{\ell_{1}}^*|+|f_{\ell_{1}}^*-f_{\ell_{2}}^*|+\cdots+|f_{\ell_{L-1}}^*-f_j^*|\leq L\Delta.\label{neighbor-L-dis}
\end{align}
Since the noise variables $\eps_i$ are assumed to be uncorrelated, centered, and have a variance of one, we can derive using \eqref{neighbor-L-dis} that for any $i=1,\ldots,n$,
\begin{align}
    \E\cro{|\overline f_{L,i}-f_i^*|^2}&=\E\cro{\left|\frac{1}{|\cN^{L}(i)|}\left(\sum_{j \in \cN^{L}(i)} Y_j\right)-f_i^*\right|^2}\nonumber\\
    &=\E\cro{\left|\frac{1}{|\cN^{L}(i)|}\left(\sum_{j \in \cN^{L}(i)}(f_j^{*}+\eps_{j})\right)-f_i^*\right|^2}\nonumber\\
    &\leq\frac{\sum_{j \in \cN^{L}(i)}|f_j^{*}-f_i^*|^2}{|\cN^{L}(i)|}+\E\cro{\left|\frac{1}{|\cN^{L}(i)|}\left(\sum_{j\in\cN^{L}(i)}\eps_{j}\right)\right|^2}\nonumber\\
    &\leq L^2\Delta^2+\frac{1}{|\cN^{L}(i)|}.\label{i-mse}
\end{align}
Summing \eqref{i-mse} for all $i=1,\ldots,n$ finally yields
\begin{align*}
\frac{1}{n}\E\cro{\|\mathbf{\overline f}_L-\mathbf{f^{*}}\|_{2}^{2}}=\frac{1}{n}\E\cro{\sum_{i=1}^{n}|\overline f_{L,i}-f_i^*|^2}\leq (L\Delta)^2+\frac{1}{n}\sum_{i=1}^{n}\frac{1}{|\cN^{L}(i)|}.
\end{align*}

\section{DERIVATION OF ASSOCIATED WEIGHTS}\label{weights}

\subsection{Associated Weights for $S^{L}$}\label{weight-s}
Recall that we can represent $S$ as shown in \eqref{S-for}, where $a_{ij}$ for $i\not=j$ indicates whether node $i$ is connected to node $j$ by an edge in $\mathcal{E}$, and $a_{ii}=1$ since self-loops are included. Similarly, we define $a_{(i\to\ell_{1}\to\ldots\to\ell_{L-1}\to j)}:=1$, if there exists a path with self-loops of length $L$ from node $i$ to node $j$ via nodes $\ell_{1},\ldots\ell_{L-1}\in\{1,\ldots,n\}$; otherwise $a_{(i\to\ell_{1}\to\ldots\to\ell_{L-1}\to j)}:=0$. In particular, when $L=1$, we have $a_{ij}=a_{(i\to j)}$. 

Let $s^{L}_{ij}$ be the entry of the matrix $S^L$ in the $i$-th row and $j$-th column. In what follows, we show by induction that for any $i,j=1,\ldots,n$,
\begin{equation}\label{s-entries}
s^{L}_{ij}=\sum_{\ell_{1},\ldots,\ell_{L-1}=1}^{n}\frac{a_{(i\to\ell_{1}\to\ldots\to\ell_{L-1}\to j)}}{(d_i+1)(d_{\ell_1}+1)\cdot \ldots \cdot(d_{\ell_{L-1}}+1)},
\end{equation}
where $d_i,d_{\ell_1},\ldots,d_{\ell_{L-1}}$ denote the edge degrees of nodes $i,\ell_{1},\ldots,\ell_{L-1}$, respectively. The equality \eqref{s-entries} then implies that the associated weight for each path is given by
$$\omega_{i}(\ell_1,\ldots,\ell_{L-1})=\frac{1}{(d_i+1)(d_{\ell_1}+1)\cdot \ldots \cdot(d_{\ell_{L-1}}+1)}.$$
One can observe from \eqref{S-for} that the conclusion holds for $L=1$. For the induction step, assume the claim holds until $L-1$, in the sense that, for any $1\leq k\leq L-1$, and $i,j=1,\ldots,n$,
$$s^{k}_{ij}=\sum_{\ell_{1},\ldots,\ell_{k-1}=1}^{n}\frac{a_{(i\to\ell_{1}\to\ldots\to\ell_{k-1}\to j)}}{(d_i+1)(d_{\ell_1}+1)\cdot \ldots \cdot(d_{\ell_{k-1}}+1)}.$$ Then 
\begin{align*}
   s^{L}_{ij}&=\sum_{\ell_{L-1}=1}^{n}s^{L-1}_{i\ell_{L-1}}\frac{a_{\ell_{L-1}j}}{d_{\ell_{L-1}}+1}\\
   &=\sum_{\ell_{L-1}=1}^{n}\cro{\left(\sum_{\ell_{1},\ldots,\ell_{L-2}=1}^{n}\frac{a_{(i\to\ell_{1}\to\ldots\to\ell_{L-2}\to\ell_{L-1})}}{(d_i+1)(d_{\ell_1}+1)\cdot \ldots \cdot(d_{\ell_{L-2}}+1)}\right)\frac{a_{(\ell_{L-1}\to j)}}{d_{\ell_{L-1}}+1}}\\
   &=\sum_{\ell_{1},\ldots,\ell_{L-1}=1}^{n}\frac{a_{(i\to\ell_{1}\to\ldots\to\ell_{L-1}\to j)}}{(d_i+1)(d_{\ell_1}+1)\cdot \ldots \cdot(d_{\ell_{L-1}}+1)},
\end{align*}
which completes the argument.

\subsection{Associated Weights for $T^{L}$}
By definition, $T=\tilde D^{-1/2}\tilde A\tilde D^{-1/2}$ and $S=\tilde D^{-1}\tilde A$. Thus, for any $L=1,2,\ldots,$
\begin{equation}\label{S-T-connection}
T^L=\tilde D^{1/2}S^{L}\tilde D^{-1/2}.
\end{equation}
For any $i,j=1,\ldots,n$, let $t^{L}_{ij}$ represent the entry of the matrix $T^L$ in the $i$-th row and $j$-th column. Building on \eqref{S-T-connection} and the formula \eqref{s-entries} for the entries of $S^{L}$, we can derive that
\begin{align*}
  t^{L}_{ij}= \frac{\sqrt{d_i+1}}{\sqrt{d_j+1}} s^{L}_{ij}&=\sum_{\ell_{1},\ldots,\ell_{L-1}=1}^{n}\frac{\sqrt{d_i+1}}{\sqrt{d_j+1}}\frac{a_{(i\to\ell_{1}\to\ldots\to\ell_{L-1}\to j)}}{(d_i+1)(d_{\ell_1}+1)\cdot \ldots \cdot(d_{\ell_{L-1}}+1)},\\
  &=\sum_{\ell_{1},\ldots,\ell_{L-1}=1}^{n}a_{(i\to\ell_{1}\to\ldots\to\ell_{L-1}\to j)}\widetilde \omega_{ij}(\ell_{1},\ldots,\ell_{L-1}),
\end{align*}
which implies that the associated weight for each path $(i\to\ell_{1}\to\ldots\to\ell_{L-1}\to j)$ is $\widetilde \omega_{ij}(\ell_{1},\ldots,\ell_{L-1})$.
\section{PROOFS OF PROPOSITIONS}
\subsection{Proof of Proposition \ref{tree-structure}}
We examine walks on the augmented graph $\mathcal{G}'$ that includes self-loops. Specifically, for an integer $r$ satisfying $0\leq r\leq L$, we consider the paths on $\mathcal{G}'$ that start at node $i$ and arrive after $L$ moves at node $j\in\cN^r(i)\setminus \cN^{r-1}(i)$. For any $j\in\cN^r(i)\setminus \cN^{r-1}(i)$, we denote the set of paths by $\mathcal{P}_{L}^{r}(i\to j)$. 

To derive an upper bound on the cardinality of $\mathcal{P}_{L}^{r}(i\to j)$, we refine the analysis as follows. For each $r$, define the set 
$$\mathcal{S}(r):=\{s\in\Z:\;r\leq s\leq L\ \text{and}\ s \equiv r\ (\text{mod}\ 2)\},$$ which represents the set of integers $s$ satisfying $r\leq s\leq L$ and having the same parity as $r$. Each path in $\mathcal{P}_{L}^{r}(i\to j)$ can then be expressed as starting at node $i$, moving $(k+r)/2$ times toward $j$, $(k-r)/2$ times away from $j$, and remaining at the same node for $L-k$ steps, for some $k\in\mathcal{S}(r)$. For any $r$ with $0\leq r\leq L$ and $k\in\mathcal{S}(r)$, let $\mathcal{P}_L^{k,r}(i \to j)$ denote all paths from node $i$ to node $j$ of length $L$ in the self-loop augmented graph $\mathcal{G}'$, characterized by $(k+r)/2$ movements toward $j$, $(k-r)/2$ movements away from $j$, and $L-k$ steps remaining at the same node. Since $\mathcal{P}_{L}^{r}(i\to j)\subseteq\cup_{k\in\mathcal{S}(r)}\mathcal{P}_{L}^{k,r}(i\to j)$, we now focus our attention on the set $\mathcal{P}_L^{k,r}(i \to j)$.

We will use $F$ to represent a  movement towards $j$, $B$ to represent a movement away from $j$, and write $R$ if the walk remains at the same node. Then each path in $\mathcal{P}_L^{k,r}(i \to j)$ corresponds to a sequence of length $L$ composed of $F$, $B$, and $R$, with $F,B$, and $R$ occurring $(k+r)/2$, $(k-r)/2$, and $L-k$ times, respectively. An  example of such a sequence is
\begin{equation}\label{exa-seq}
\underbrace{R,\cdots, R,}_{\text{($L-k$)-times}}\underbrace{F,\cdots, F,}_{\text{$r$-times}}\underbrace{B,\cdots, B,}_{\text{$(k-r)/2$-times}}\underbrace{F,\cdots, F}_{\text{$(k-r)/2$-times}}.
\end{equation}
Under the locally rooted tree structure assumption on $\mathcal{G}$, there is a unique sequence of edges $(i\to\ell_{1}\to\ldots\to\ell_{r-1}\to j)$ connecting node $i$ to node $j\in\cN^{r}(i)\setminus\cN^{r-1}(i)$ in the original graph. All paths in $\mathcal{P}_L^{k,r}(i \to j)$ in $\mathcal{G}'$ are derived from variations of this path $(i\to\ell_{1}\to\ldots\to\ell_{r-1}\to j)$ in $\mathcal{G}$. For the $F$ and $R$ steps, the resulting node after one movement is completely determined, ensuring that multiple branches will 
not arise based on the existing path before employing either $F$ or $R$ in $\mathcal{G}'$. In contrast, in every $B$ step, we have $d-1$ choices to move to another node. Focusing on the  $(k-r)/2$ occurrences of $B,$ we can write the walk in the form
\begin{equation}\label{A-seq-format}
\ldots,\underbrace{B,\cdots, B,}_{\text{$q_1$-times}}\ldots,\underbrace{B,\cdots,B,}_{\text{$q_2$-times}}\ldots,\underbrace{B,\cdots, B,}_{\text{$q_3$-times}}\ldots,\underbrace{B,\cdots,B}_{\text{$q_m$-times}},\ldots,
\end{equation}
with $q_1,\ldots,q_m\geq1$ and $q_1+q_2+\ldots+q_m=(k-r)/2$. The number of paths  of the form \eqref{A-seq-format} in the augmented graph is hence bounded by $d^{q_1}d^{q_2}\ldots d^{q_{m}}=d^{(k-r)/2}.$ In a path we can in each step choose one of the three options $F,B,R.$ Thus there are at most $3^L$ different ways in \eqref{A-seq-format} and for any $0\leq r\leq L$, $j\in \cN^r(i)\setminus \cN^{r-1}(i)$, and $k\in\mathcal{S}(r)$,
\begin{equation}\label{path-rk}
    \left|\mathcal{P}_L^{k,r}(i \to j)\right|\leq
    3^Ld^{(k-r)/2}.
\end{equation}

Moreover, observe that for any $0\leq r\leq L,$ due to the locally rooted tree structure of $\mathcal{G}$, we have
\begin{equation}\label{rooted-tree-bound}
    |\cN^r(i)\setminus\cN^{r-1}(i)|= d^r, 
\end{equation}
with the convention that $\cN^0(i)=\{i\}$ and $\cN^{-1}(i)=\emptyset$. Therefore, using \eqref{path-rk} and \eqref{rooted-tree-bound}, 
\begin{align}
\E\cro{\big(S^L \bm{\eps}\big)_i^2}
&=\sum_{j=1}^{n}\cro{\sum_{(i\to\ell_1\to\ldots\to\ell_{L-1}\to j)\in \mathcal{P}_L(i\to j)}\omega_{i}(\ell_1,\ldots,\ell_{L-1})}^2\nonumber\\
&\leq\sum_{r=0}^{L}\sum_{j\in \cN^r(i)\setminus \cN^{r-1}(i)}\Bigg(\sum_{(i\to\ell_1\to\ldots\to\ell_{L-1}\to j) \in \mathcal{P}_L^{r}(i \to j)}(d+1)^{-L}\Bigg)^2 \nonumber\\
&\leq\sum_{r=0}^{L}\sum_{j\in \cN^r(i)\setminus \cN^{r-1}(i)}\Bigg(\sum_{k\in\mathcal{S}(r)}\sum_{(i\to\ell_1\to\ldots\to\ell_{L-1}\to j) \in \mathcal{P}_L^{k,r}(i \to j)}(d+1)^{-L}\Bigg)^2 \nonumber\\
&\leq\sum_{r=0}^{L}d^{r}\cro{\sum_{k\in\mathcal{S}(r)} \,3^Ld^{\frac{k-r}{2}} (d+1)^{-L}}^2 \nonumber\\
&\leq3^{2L}(d+1)^{-2L}\sum_{r=0}^{L}\left(\sum_{k\in\mathcal{S}(r)}d^{\frac{k}{2}}\right)^2.\label{b-prop-1}
\end{align}
For $d>1$,
\begin{align*}
\sum_{r=0}^{L}\left(\sum_{k\in\mathcal{S}(r)}d^{\frac{k}{2}}\right)^2&=\sum_{r=0}^{L}\left(\sum_{\ell=0}^{\lfloor\frac{L-r}{2}\rfloor}d^{\frac{r}{2}+\ell}\right)^2\\
&=\sum_{r=0}^{L}\cro{d^{\frac{r}{2}}\frac{(1-d^{\lfloor\frac{L-r}{2}\rfloor+1})}{1-d}}^2\\
&\leq\sum_{r=0}^{L}\frac{(d^{\frac{L}{2}+1}-d^{\frac{r}{2}})^2}{(d-1)^2}\\
&\leq(L+1)d^{L}\frac{d^2}{(d-1)^2}.
\end{align*}
Combining the previous inequalities with $d/(d-1)\leq 2$, we finally obtain 
$$\E\cro{\big(S^L \bm{\eps}\big)_i^2}\leq 4 \cdot 3^{2L} (d+1)^{-L}(L+1) .$$
\subsection{Proof of Proposition \ref{tree-structure-slow}}
In this case, we only need to focus on the single path $(i\to\ell_1\to\ldots\to\ell_{L-1}\to j)$, where $d_{\ell_1},\ldots,d_{\ell_{L-1}}\leq3$ and $d_i=d$ under the given condition. For this single path, the associated weight is 
\begin{equation}\label{single-path-weight}
    \omega_{i}(\ell_1,\ldots,\ell_{L-1})=\frac{1}{(d_i+1)(d_{\ell_1}+1)\cdot\ldots\cdot(d_{\ell_{L-1}}+1)}\geq\frac{4^{1-L}}{d+1}.
\end{equation}
A direct calculation using \eqref{single-path-weight} gives the result, 
\begin{align*}
    \E\big[\big(S^L \bm{\eps}\big)_i^2\big]&=
\sum_{j=1}^{n}\cro{\sum_{(i\to\ell_1\to\ldots\to\ell_{L-1}\to j)\in \mathcal{P}_L(i\to j)}\omega_{i}(\ell_1,\ldots,\ell_{L-1})}^2\\
&\geq\cro{\frac{1}{(d+1)\cdot 4^{L-1}}}^2\\
&=(d+1)^{-2}4^{2-2L}.
\end{align*}

\subsection{Proof of Proposition \ref{circle-lem}}
Consider all the nodes on the cycle $\mathcal{C}$, where the nodes are reindexed from 1 to $r$, starting from the node $i=\mathcal{C}\cap\mathcal{H}$. It is sufficient to consider only paths that remain in the cycle $\mathcal{C}$, move in the first step to either node $2$ or node $r$, and do not visit node $1$ again. Since self-loops are allowed, walking along these paths gives in every step at least two possible nodes to move to in the next step. The total number of paths is thus $\geq 2^L.$ Then by the pigeonhole principle, there exists at least one node $j\in\{2,\ldots,r\}$ such that $|\mathcal{P}_L(1 \to j)|\geq2^L/(r-1).$ Thus,
\begin{align*}
\E\big[\big(S^L \bm{\eps}\big)_{1}^2\big]
&\geq\Bigg(\sum_{(i\to\ell_1\to \ldots\to\ell_{L-1}\to j) \in \mathcal{P}_L(1 \to j)} \, \frac{1}{(d_i+1)3^{L-1}}\Bigg)^2\\
&\geq\cro{\frac{2^L}{r-1}\frac{1}{(d_i+1)3^{L-1}}}^2\\
&=\cro{\frac{3}{(d_i+1)(r-1)}}^{2}1.5^{-2L}.
\end{align*}

\section{ADDITIONAL PROPOSITIONS AND PROOFS FOR $P=T^L$}
In this section, we will repeatedly use the equality 
\begin{equation}\label{s-t-conn-weight}
    \widetilde \omega_{ij}(\ell_1,\ldots,\ell_{L-1}):=\Big(\frac{d_i+1}{d_j+1}\Big)^{1/2}\omega_{i}(\ell_1,\ldots,\ell_{L-1})
\end{equation}
derived in the main part of the paper.

\begin{customthm}{1'}
Assume the local graph around node $i$ is a rooted tree, in the sense that the subgraph induced by the vertices $\cN^L(i)$ forms a  rooted tree with root $i$ and all nodes having edge degree $d$. Then, we have 
\begin{align*}
    \E\big[\big(T^L \bm{\eps}\big)_i^2\big]\leq 4(d+1)^{-L} (L+1)3^{2L}.
\end{align*}
\end{customthm}
\begin{proof}
Using \eqref{s-t-conn-weight},
\begin{align*}
\E\cro{\big(T^L \bm{\eps}\big)_i^2}
&=\sum_{j=1}^{n}\cro{\sum_{(i\to\ell_1\to\ldots\to\ell_{L-1}\to j)\in \mathcal{P}_L(i\to j)}\widetilde \omega_{ij}(\ell_1,\ldots,\ell_{L-1})}^2\\
&=\sum_{j=1}^{n}\cro{\sum_{(i\to\ell_1\to\ldots\to\ell_{L-1}\to j)\in \mathcal{P}_L(i\to j)}\Big(\frac{d_i+1}{d_j+1}\Big)^{1/2}\omega_{i}(\ell_1,\ldots,\ell_{L-1})}^2\\
&=\sum_{j=1}^{n}\cro{\sum_{(i\to\ell_1\to\ldots\to\ell_{L-1}\to j)\in \mathcal{P}_L(i\to j)}\omega_{i}(\ell_1,\ldots,\ell_{L-1})}^2\\
&=\E\cro{\big(S^L \bm{\eps}\big)_i^2}\\
&\leq4(d+1)^{-L} (L+1)3^{2L},
\end{align*}
where the second-to-last equality is due to $d_i=d_j=d$, and the last inequality follows from Proposition~\ref{tree-structure} in the paper.
\end{proof}

\begin{customthm}{2'}
If there exists a path from node $i$ to node $j$ via nodes $\ell_1,\ldots,\ell_{L-1}\in\{1,\ldots,n\}$ such that $d_{\ell_1}, \ldots,  d_{\ell_{L-1}}\leq 3,$ then 
\begin{align*}
    \E\big[\big(T^L \bm{\eps}\big)_i^2\big]
    \geq \frac{4^{2-2L}}{(d+1)(d_j+1)},
\end{align*}
where $d_{j}$ denotes the edge degree of node $j$.
\end{customthm}
\begin{proof}
Considering the single path $(i\to\ell_{1}\to\ldots\to\ell_{L-1}\to j)$ with $d_{\ell_1},\ldots,d_{\ell_{L-1}}\leq3$ and $d_i=d$, we obtain by applying \eqref{s-t-conn-weight} that
\begin{align*}
    \E\big[\big(T^L \bm{\eps}\big)_i^2\big]&=\sum_{j=1}^{n}\cro{\sum_{(i\to\ell_1\to\ldots\to\ell_{L-1}\to j)\in \mathcal{P}_L(i\to j)}\Big(\frac{d_i+1}{d_j+1}\Big)^{1/2}\omega_{i}(\ell_1,\ldots,\ell_{L-1})}^2\\
    &\geq\cro{\Big(\frac{d+1}{d_j+1}\Big)^{1/2}\frac{1}{(d+1)\cdot 4^{L-1}}}^2\\
    &=\frac{4^{2-2L}}{(d+1)(d_j+1)}.
\end{align*}
\end{proof}

\begin{customthm}{3'}
Assume that the graph decomposes into a cycle $\mathcal{C}$ of length $r$ and a graph $\mathcal{H}$ in the sense that $\mathcal{C}$ and $\mathcal{H}$ share exactly one node $i:=\mathcal{C}\cap\mathcal{H}$ and there are no edges connecting $\mathcal{C}\setminus\{i\}$ and $\mathcal{H}\setminus\{i\}.$ Then, for any $L=1,2,\ldots$,
\begin{equation*}
\E\big[\big(T^L \bm{\eps}\big)_i^2\big]\geq \frac{3}{(d_i+1)(r-1)^2}1.5^{-2L}.
\end{equation*}
\end{customthm}
\begin{proof}
Following the proof of Proposition 3, in which the nodes on the cycle $\mathcal{C}$ have been reindexed from 1 to $r$, starting from $i=\mathcal{C}\cap\mathcal{H}$, there exists at least one node $j\in\{2,\ldots,r\}$ such that $|\mathcal{P}_L(1 \to j)|\geq2^L/(r-1).$ Using \eqref{s-t-conn-weight}, we can derive that
\begin{align*}
\E\big[\big(T^L \bm{\eps}\big)_i^2\big] &=\sum_{j=1}^{n}\cro{\sum_{(i\to\ell_1\to\ldots\to\ell_{L-1}\to j)\in \mathcal{P}_L(i\to j)}\Big(\frac{d_i+1}{d_j+1}\Big)^{1/2}\omega_{i}(\ell_1,\ldots,\ell_{L-1})}^2\\
&\geq\cro{\sum_{(i\to\ell_1\to\ldots\to\ell_{L-1}\to j)\in \mathcal{P}_L(i\to j)}\Big(\frac{d_i+1}{d_j+1}\Big)^{1/2}\omega_{i}(\ell_1,\ldots,\ell_{L-1})}^2\\
&\geq\cro{\frac{2^L}{r-1}\Big(\frac{d_i+1}{3}\Big)^{1/2}\frac{1}{(d_i+1)3^{L-1}}}^2\\
&=\frac{3}{(d_i+1)(r-1)^2}1.5^{-2L}.
\end{align*}
\end{proof}

\section{EXPERIMENTS}
\subsection{Real Data}
\begin{figure}[!ht]
    \centering
\includegraphics[width=\linewidth]{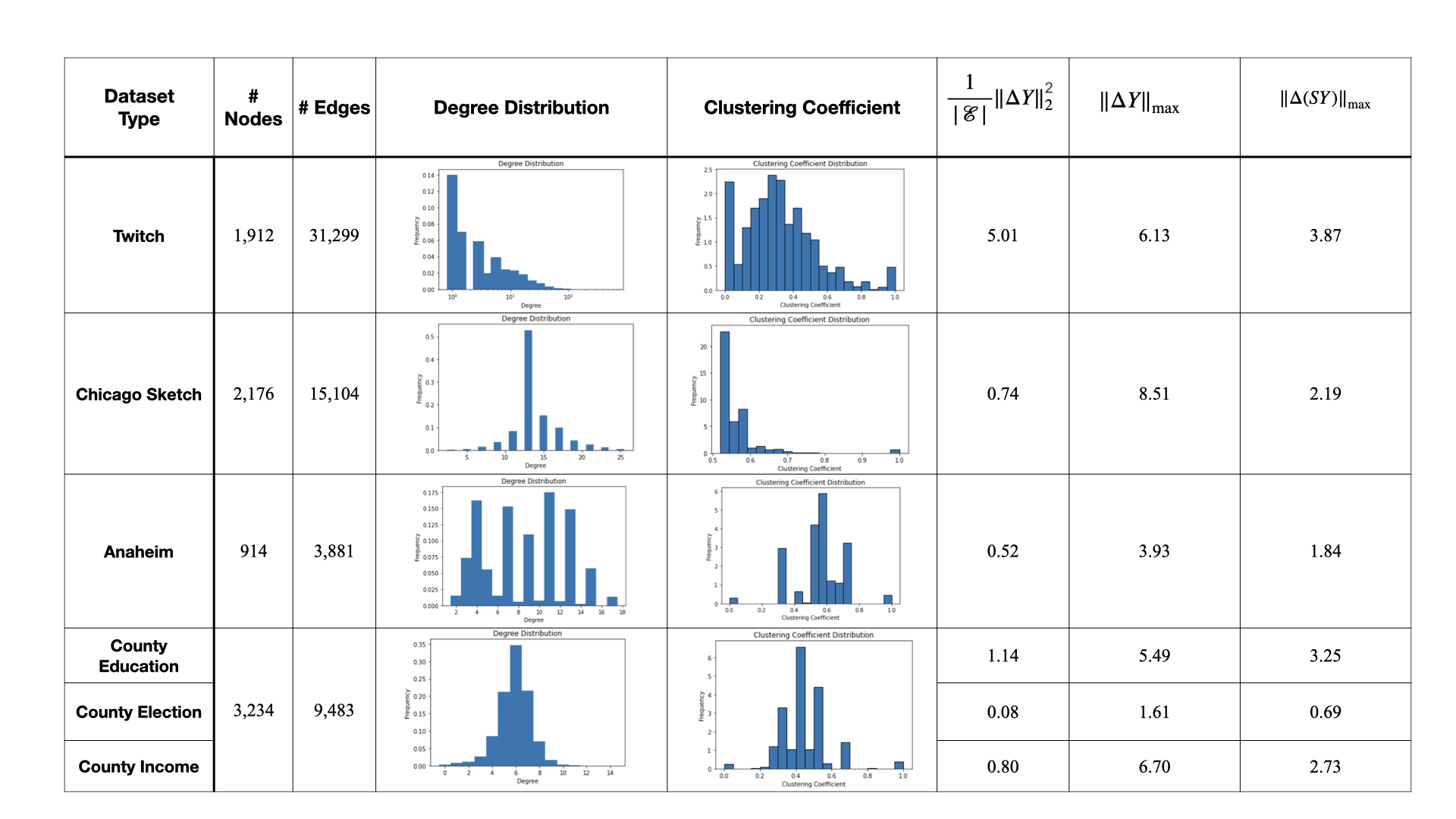}
    \caption{Properties of the datasets used in the real-data experiment section. The last 3 columns measure the signal roughness, defined either in terms of the $\ell_2$ norm ($\| \Delta  {\bf Y}\|^2_2/|\mathcal{E}| = [\sum_{(i,j) \in \mathcal{E}} (Y_i-Y_j)^2]/|\mathcal{E}| $) or the $\ell_{\infty}$ norm ($\| \Delta  {\bf Y}\|_{\max} = \max_{(i,j) \in \mathcal{E}} |Y_i-Y_j| $). The last column represent the graph smoothness over the neighborhood (after one convolution $S$).}
    \label{fig:enter-label}
\end{figure}

\begin{figure}[!ht]
    \centering
\includegraphics[width=\linewidth]{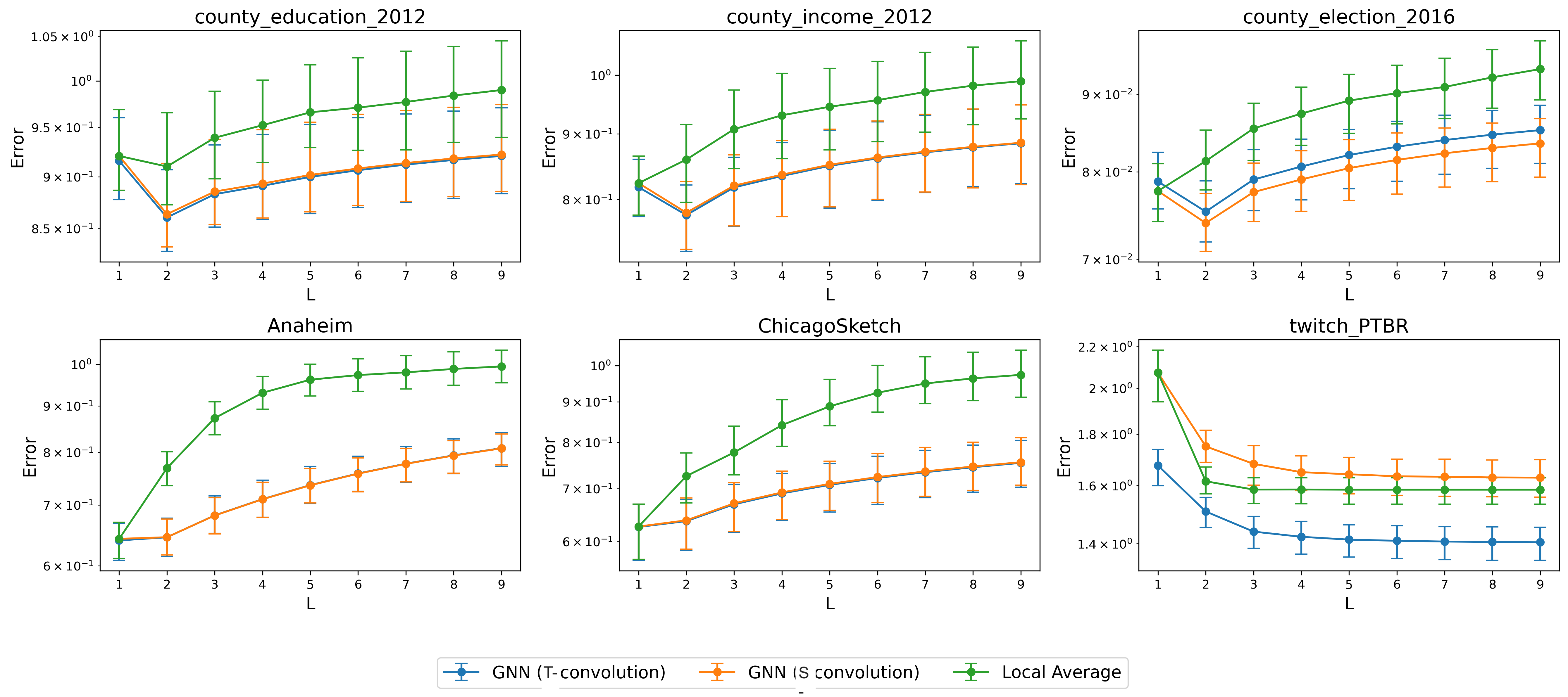}
    \caption{Results of the denoising experiment on real data across different regression datasets.}
\label{fig:denoising_real}
\end{figure}

In this subsection, we present some of the properties of the datasets used in the main text of this paper. Figure~\ref{fig:enter-label} shows the differences in terms of graph topologies and graph signal (assessed in terms of their ``roughness''). In particular, we note that the county graphs are extremely regular in structure, while the Twitch dataset exhibits a heavier tail in terms of degree distribution. Among the county graphs, the county election data is the smoothest. Interestingly, in this setting, the GCN convolution outperforms the GraphSAGE convolution in both denoising (see Figure~\ref{fig:denoising_real}) and prediction.

\subsection{Simulated Data}
\begin{figure}[!ht]
    \centering
\includegraphics[width=\linewidth]{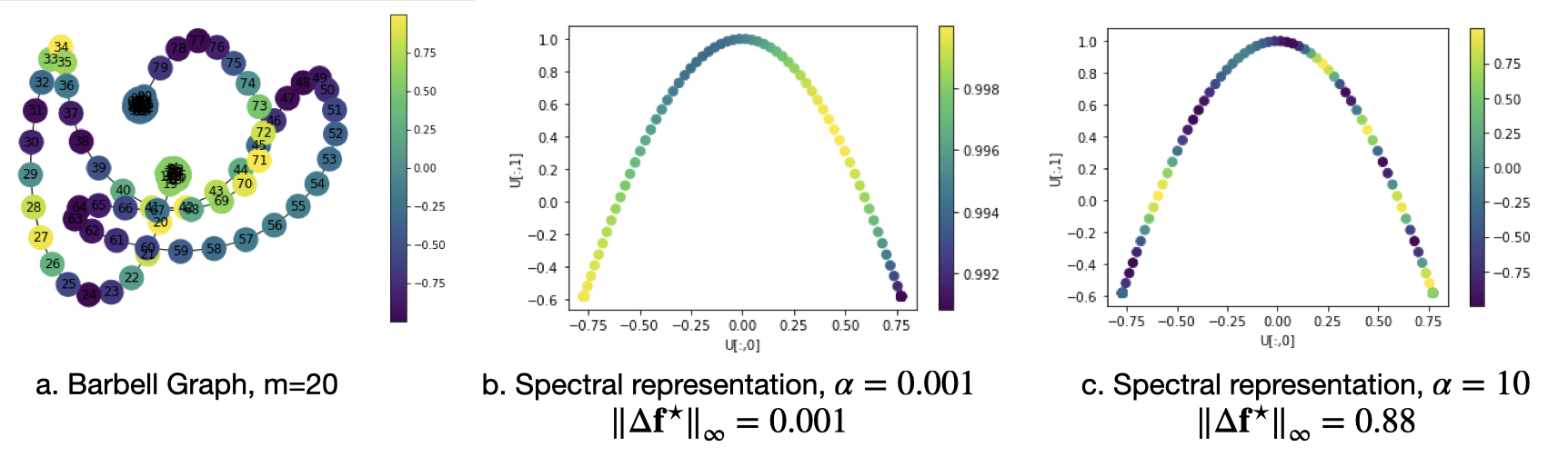}
    \caption{Barbell graph. Figure (a) shows the graph with corresponding graph signal $f_i^* = 2\cos(U_{i\cdot}\beta), \beta = (-\alpha, \alpha)^{\top}$ with $\alpha=10$. Here $U_{i\cdot}\in\mathbb{R}^{1\times2}$ represents the two dimensional spectral embedding of the graph. Figure (b) shows the spectral embedding, colored by signal for $\alpha=0.001$ and figure (c) for $\alpha=10$. }
    \label{fig:enter-label-2}
\end{figure}

\begin{figure}[!ht]
    \centering
\includegraphics[width=\linewidth]{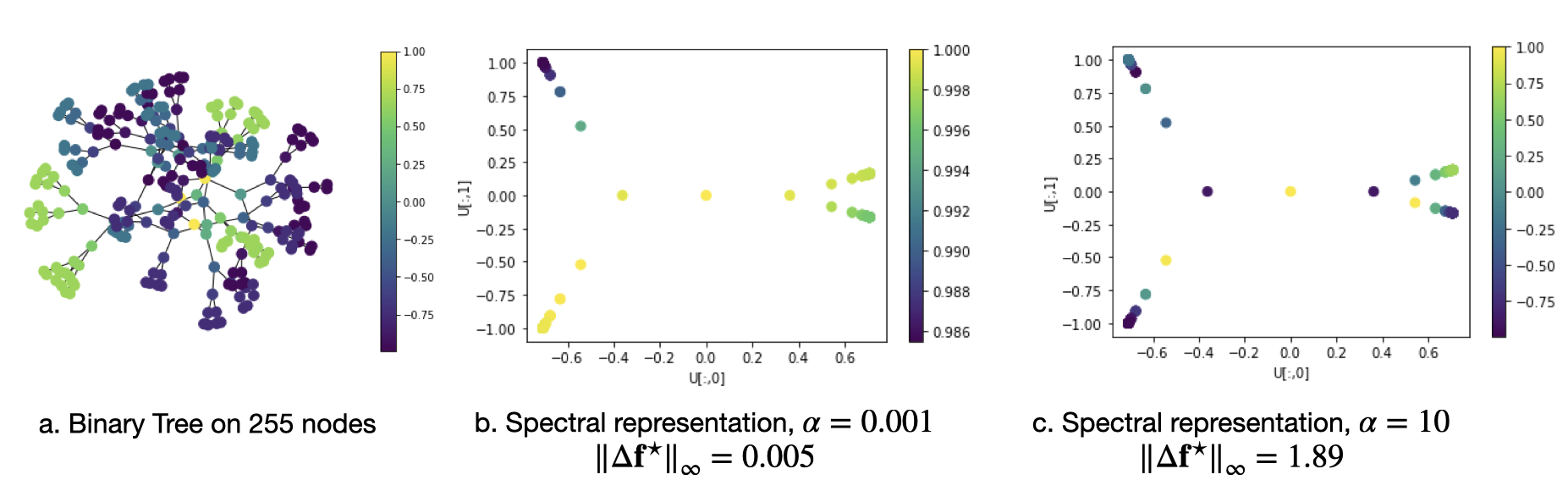}
    \caption{Binary Tree. Figure (a) shows the graph with corresponding graph signal $f_i^* = 2\cos(U_{i\cdot}\beta), \beta = (-\alpha, \alpha)^{\top}$ with $\alpha=10$. Here $U_{i\cdot}\in\mathbb{R}^{1\times2}$ represents the two dimensional spectral embedding of the graph. Figure (b) shows the spectral embedding, colored by signal for $\alpha=0.001$ and figure (c) for $\alpha=10$.}
    \label{fig:enter-label-3}
\end{figure}

\begin{figure}[!ht]
    \centering
\includegraphics[width=\linewidth]{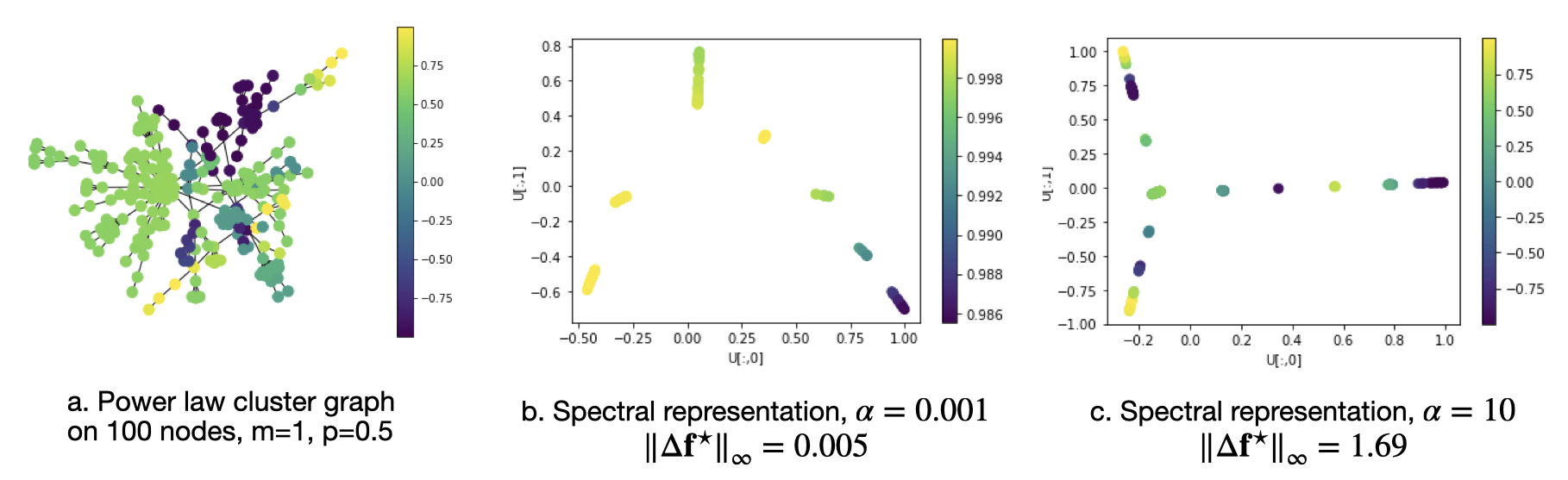}
    \caption{Power Law Cluster graph. Figure (a) shows the graph with corresponding graph signal $f_i^* = 2\cos(U_{i\cdot}\beta), \beta = (-\alpha, \alpha)^{\top}$ with $\alpha=10$. Here $U_{i\cdot}\in\mathbb{R}^{1\times2}$ represents the two dimensional spectral embedding of the graph. Figure (b) shows the spectral embedding, colored by signal for $\alpha=0.001$ and figure (c)  for $\alpha=10$. }
    \label{fig:enter-label-4}
\end{figure}

\begin{figure}[!ht]
    \centering
\includegraphics[width=\linewidth]{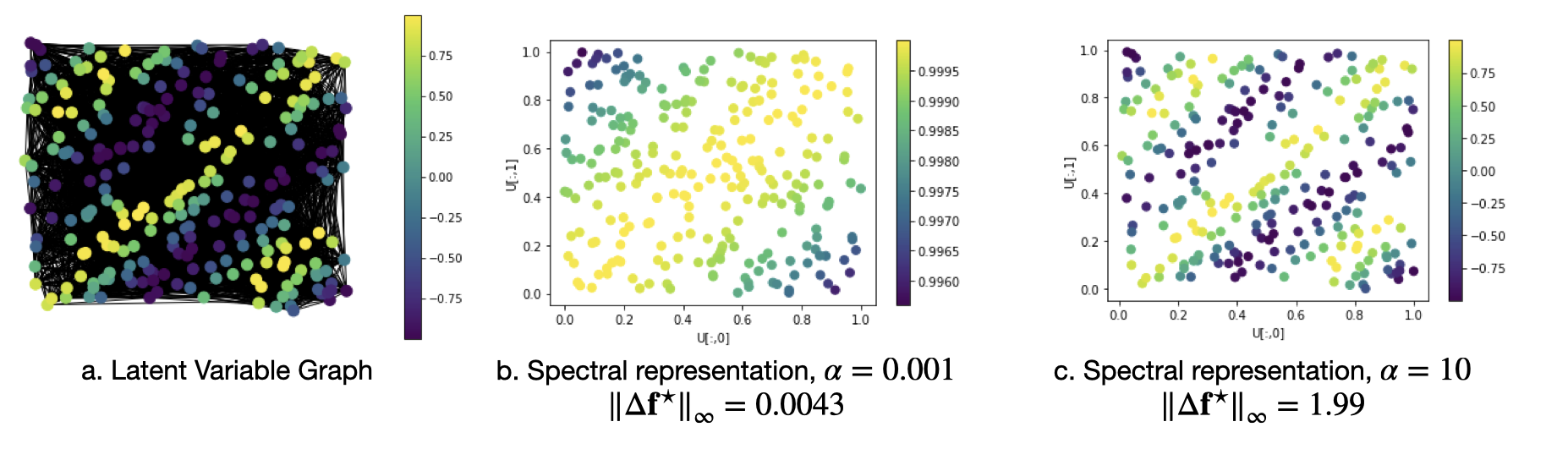}
    \caption{Latent Variable graph. Figure (a) shows the graph with corresponding graph signal $f_i^* = 2\cos(U_{i\cdot}\beta), \beta = (-\alpha, \alpha)^{\top}$ with $\alpha=10$. Here $U_{i\cdot}\in\mathbb{R}^{1\times2}$ represents the two dimensional spectral embedding of the graph. Figure (b) shows the spectral embedding, colored by signal for $\alpha=0.001$ and figure (c)  for $\alpha=10$.}
    \label{fig:enter-label-5}
\end{figure}

In this subsection, we provide examples of the different graphs used in our synthetic experiments. The following four figures (Figures~\ref{fig:enter-label-2}, \ref{fig:enter-label-3}, \ref{fig:enter-label-4}, and \ref{fig:enter-label-5}) illustrate the various graph topologies and the corresponding graph signals.

\begin{figure}[!ht]
    \centering
    \includegraphics[width=\linewidth]{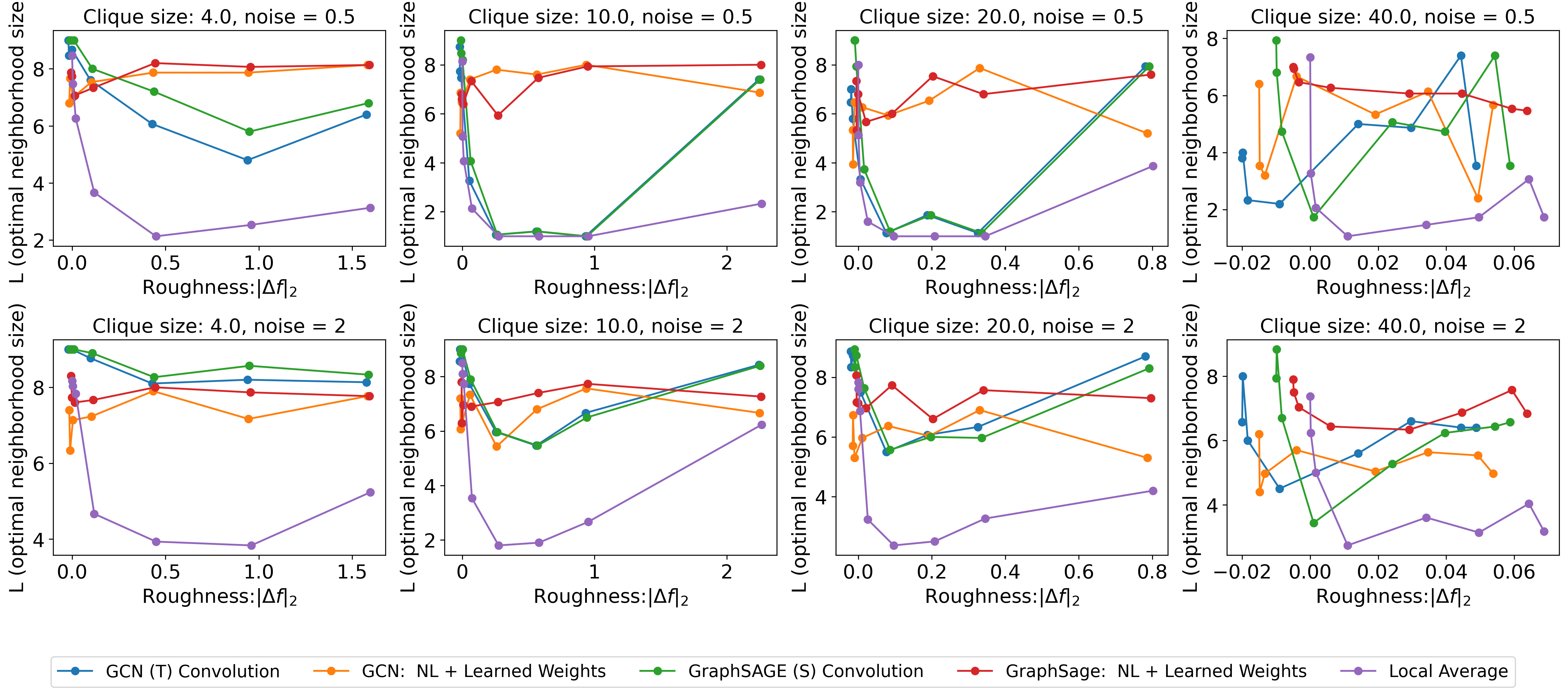}
\caption{Optimal number of convolutions as a function of the roughness of $\mathbf{f}^*$ (here defined as  $\|\Delta {\bf f^*}\|_2 =  \sqrt{[\sum_{ (i,j) \in \mathcal{E}} (f_i^{*} - f_j^{*})^2]/|\mathcal{E}|}$) on the Barbell graph. Each column corresponds to a different clique size $m$, (with, for instance, 10 meaning that the two cliques on either side of the Barbell graph are of size 10) and each row to a value of the noise $\sigma^2$.}
    \label{fig:opt_L_barbell}
\end{figure}

\begin{figure}[!ht]
    \centering
    \includegraphics[width=\linewidth]{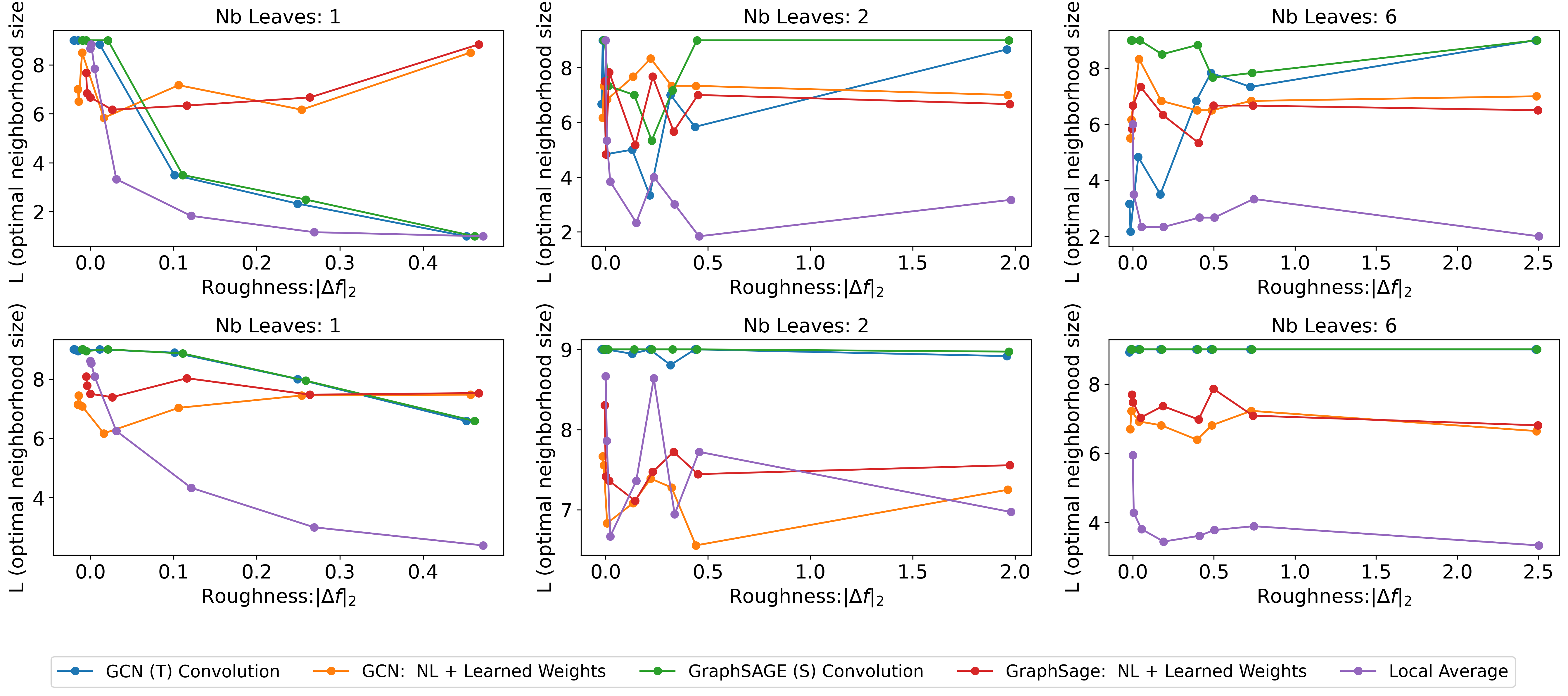}
\caption{Optimal number of convolutions as a function of the roughness of $\mathbf{f}^*$ (here defined as  $\|\Delta {\bf f^*}\|_2 =  \sqrt{[\sum_{ (i,j) \in \mathcal{E}} (f_i^{*} - f_j^{*})^2]/|\mathcal{E}|}$) on a tree graph. Each column corresponds to a different branching number (or number of leaves) $k$,  and each row to a value of the noise $\sigma^2$: the top row corresponds to a value of $\sigma^2=0.5$, while the bottom row corresponds to $\sigma^2=2$.}
    \label{fig:opt_L_tree}
\end{figure}

\begin{figure}[!ht]
    \centering
\includegraphics[width=\linewidth]{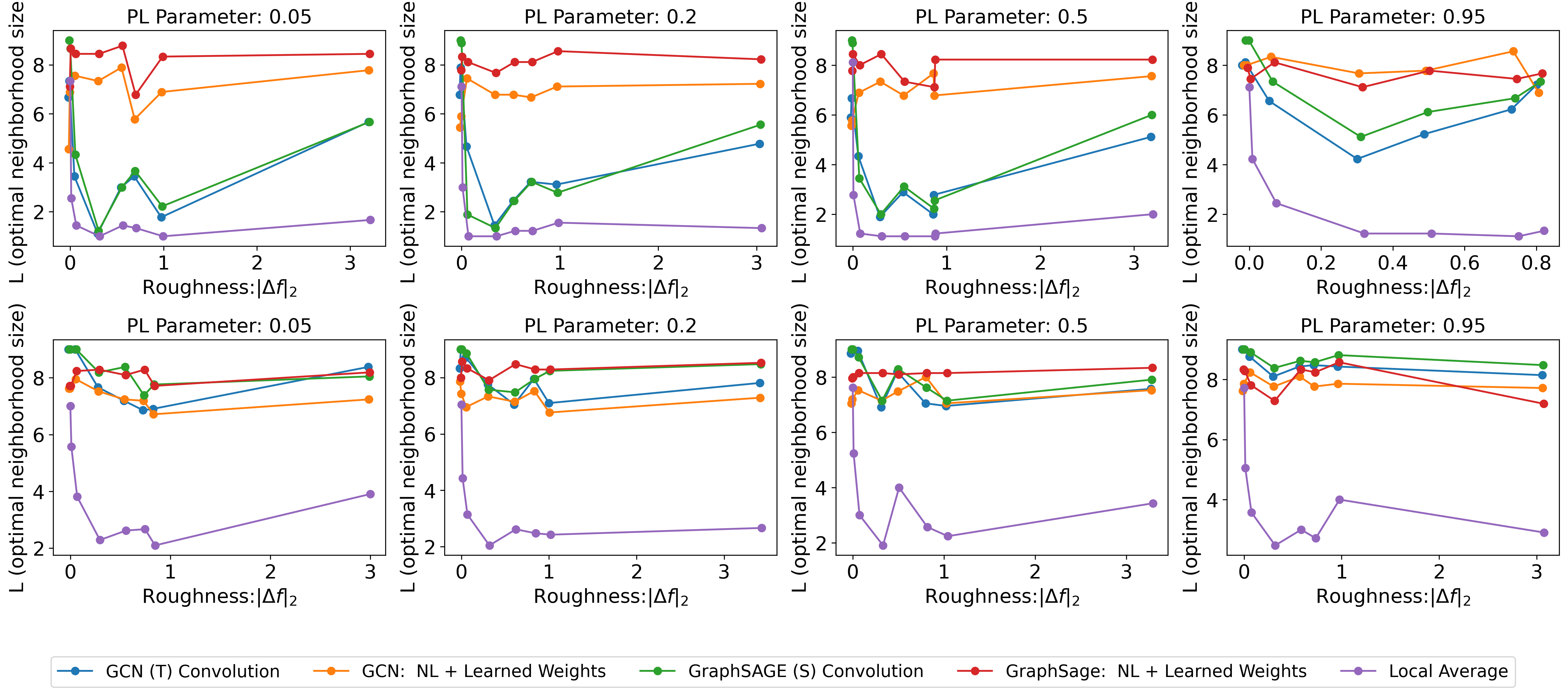}
\caption{Optimal number of convolutions as a function of the roughness of $\mathbf{f}^*$ (here defined as $\|\Delta {\bf f^*}\|_2 =  \sqrt{[\sum_{ (i,j) \in \mathcal{E}} (f_i^{*} - f_j^{*})^2]/|\mathcal{E}|}$) on the power-law cluster graph. Each column corresponds to a different clustering  parameter $p$,  and each row to a value of the noise $\sigma^2$: the top row corresponds to a value of $\sigma^2=0.5$, while the bottom row corresponds to $\sigma^2=2$.}
\label{fig:opt_L_PL}
\end{figure}

The next set of plots (Figures \ref{fig:opt_L_barbell}, \ref{fig:opt_L_tree} and  \ref{fig:opt_L_PL}) highlights the optimal number of convolutions as a function of the graph roughness. Overall, we observe the same phenomenon on the GCN and GraphSAGE convolutions as for the latent variable graph described in the main text: as the roughness of the graph increases, the optimal number of convolutions decreases. In all cases, we observe that when the neighborhood becomes too uninformative, the optimal number of layers increases again. This, however, needs to be understood alongside with the error, which increases substantially.

\begin{figure}[!ht]
    \centering
\includegraphics[width=\linewidth]{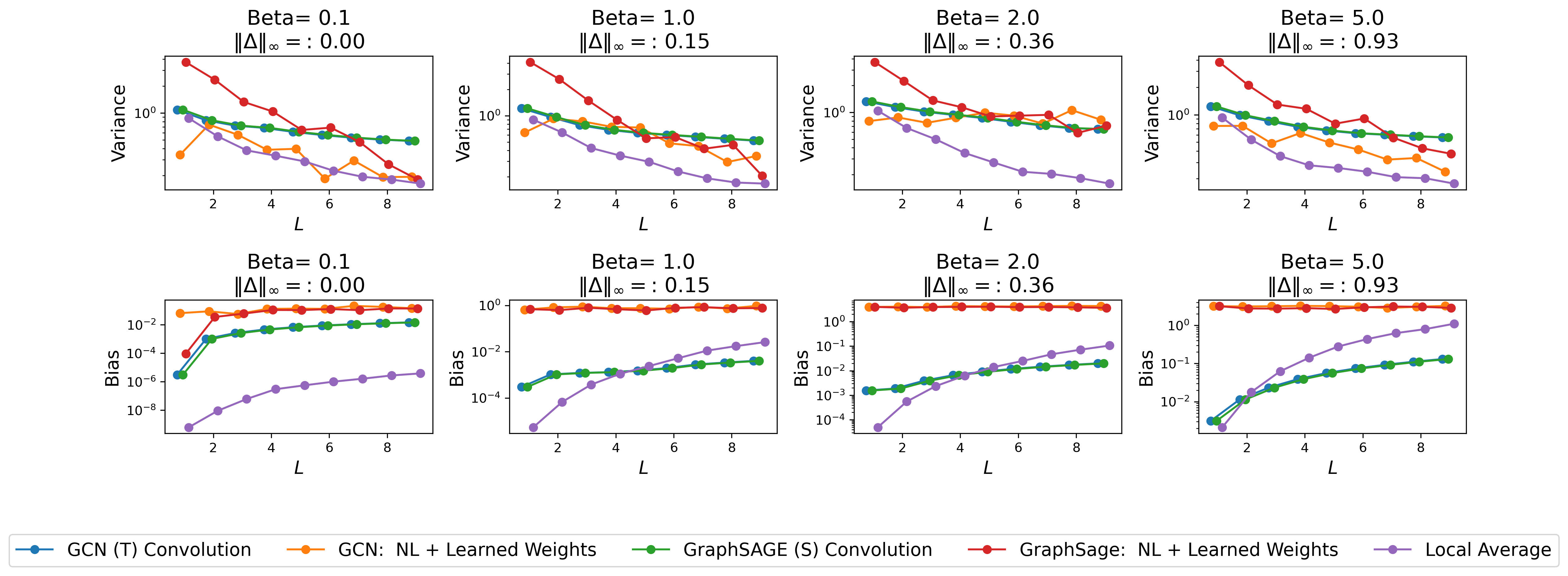}
    \caption{Bias-Variance as a function of $L$ for the Barbell graph, clique size $m=20$, $\sigma^2=2$.}
    \label{fig:barbell-20}
\end{figure}

\begin{figure}[!ht]
    \centering
\includegraphics[width=\linewidth]{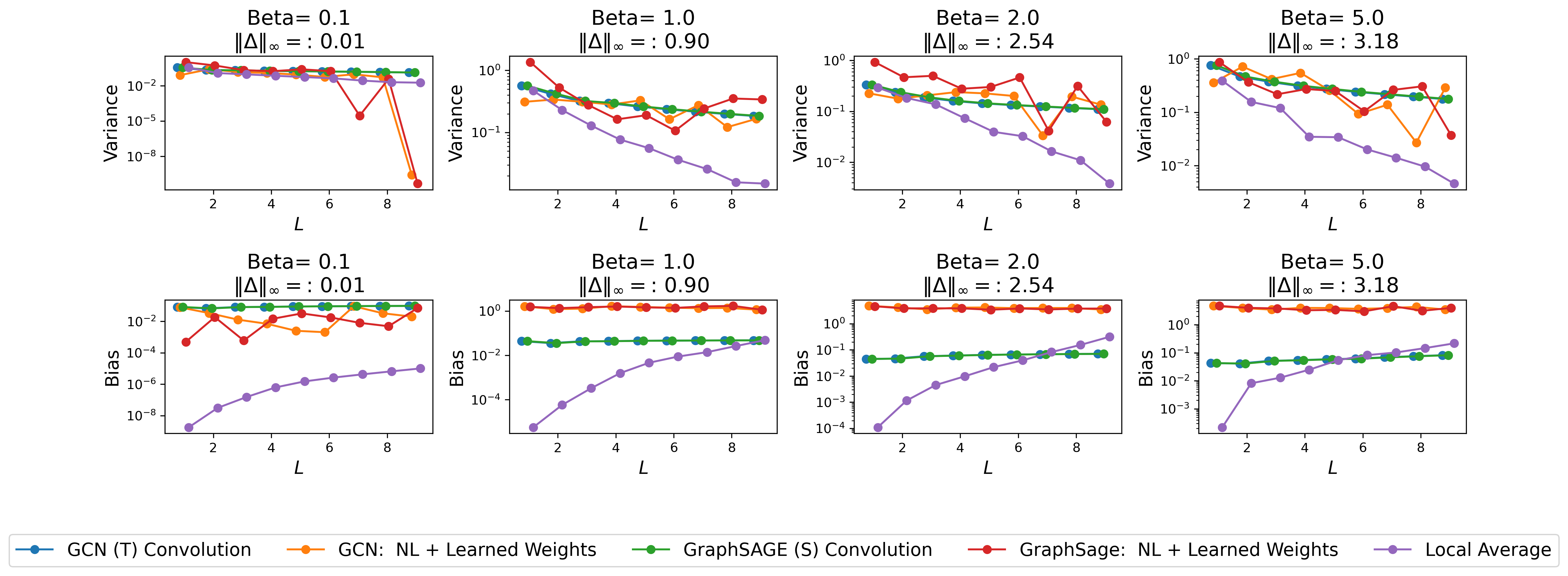}
 \caption{Bias-Variance as a function of $L$ for the tree (with branching number equal to 2), $\sigma^2=1$.}
    \label{fig:tree-2}
\end{figure}

\begin{figure}[!ht]
    \centering
\includegraphics[width=\linewidth]{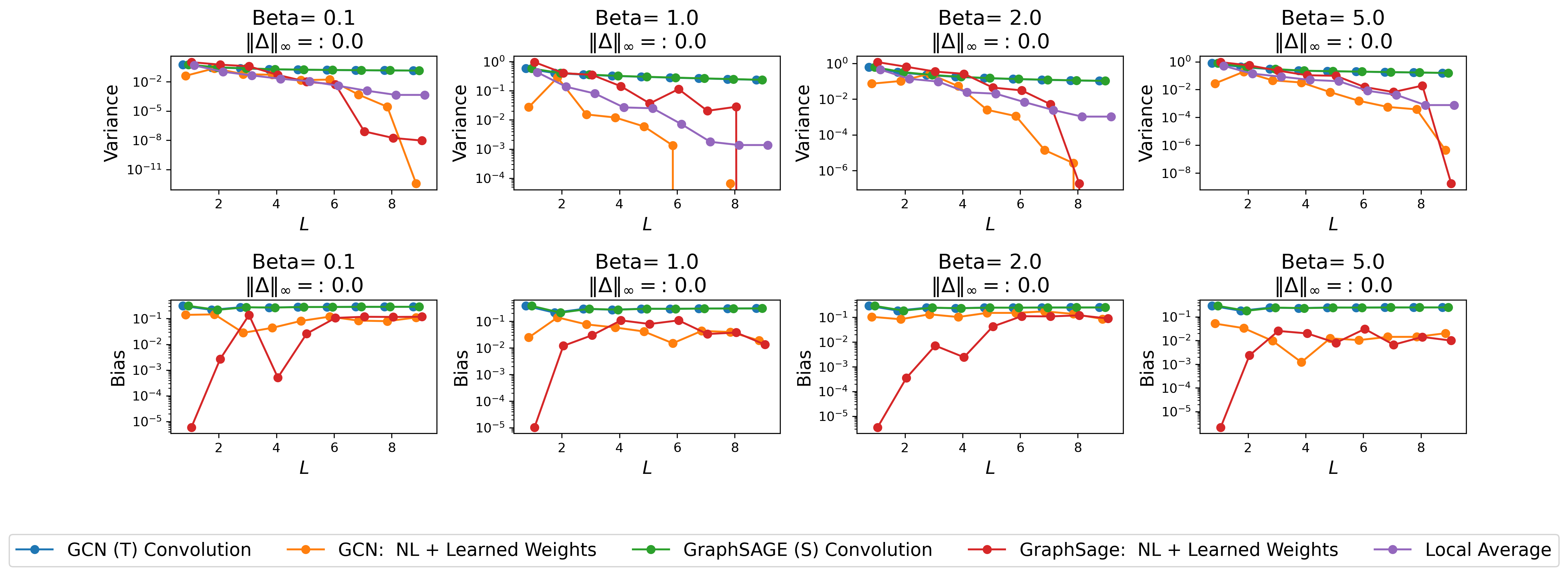}
 \caption{Bias-Variance as a function of $L$ for the tree (with branching number equal to 4), $\sigma^2=1$.}
    \label{fig:tree-4}
\end{figure}

\begin{figure}[!ht]
    \centering
\includegraphics[width=\linewidth]{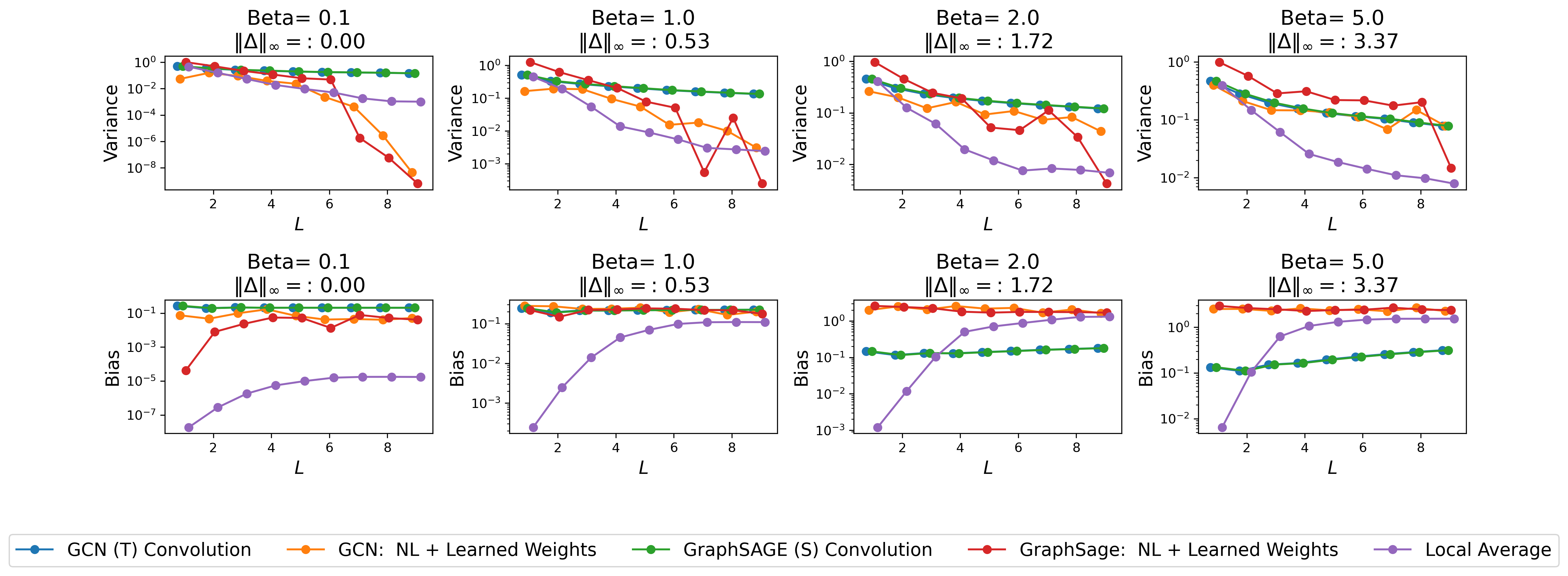}
 \caption{Bias-Variance as a function of $L$ for the power law cluster graph ($m=1, p=0.1$), $\sigma^2=1$.}
    \label{fig:PL-3}
\end{figure}

\begin{figure}[!ht]
    \centering
\includegraphics[width=\linewidth]
{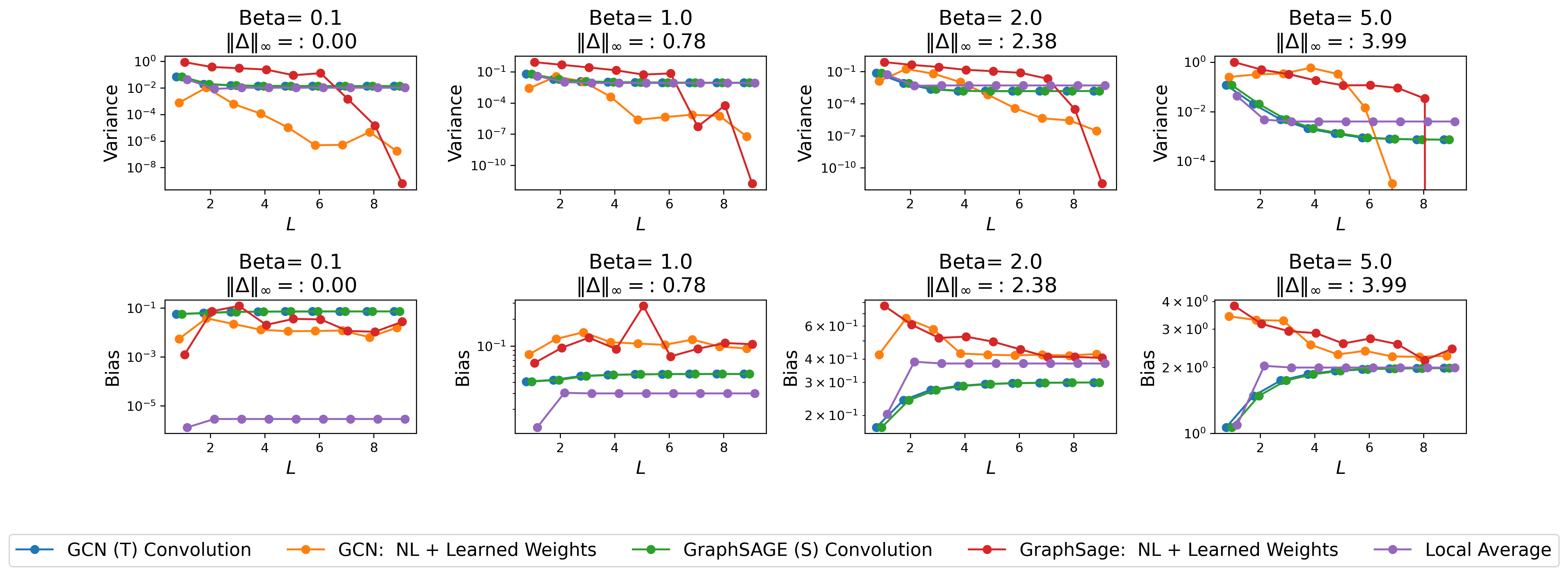}
 \caption{Bias-Variance as a function of $L$ for the latent variable graph, $\sigma^2=1$.}
    \label{fig:latent-0}
\end{figure}

\begin{figure}[!ht]
    \centering
\includegraphics[width=\linewidth]
{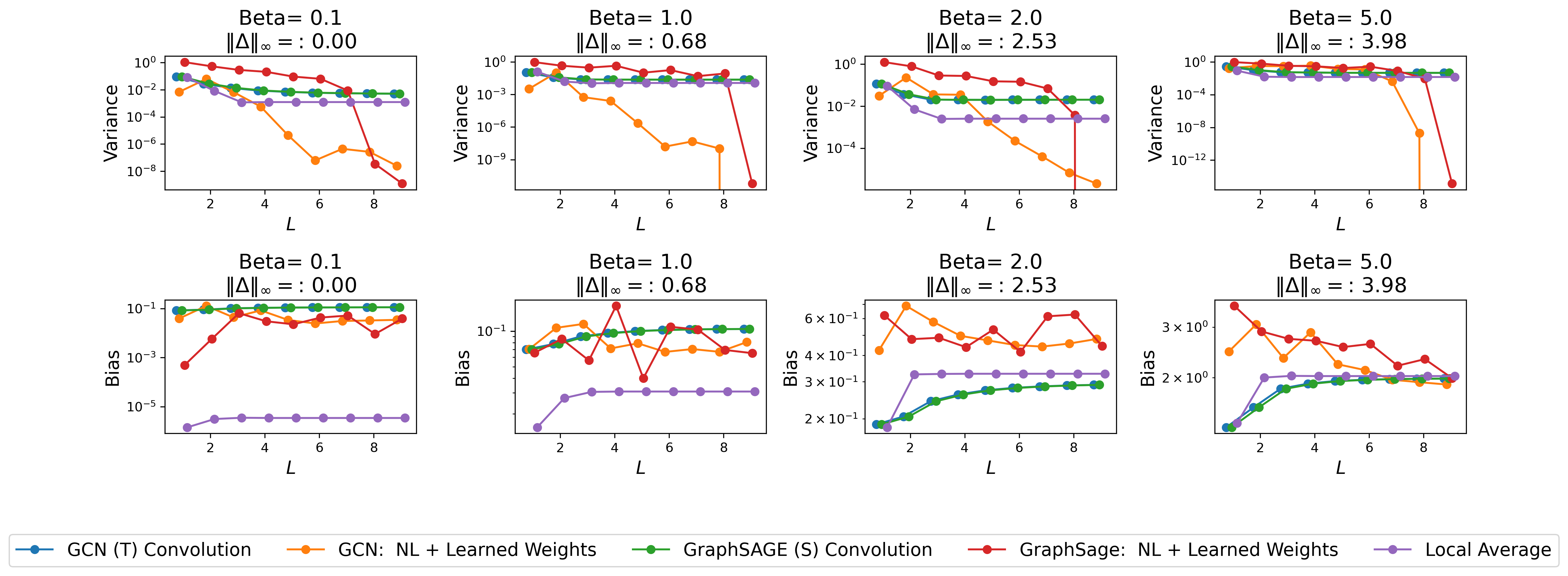}
 \caption{Bias-Variance as a function of $L$ for the sparsified latent variable graph ($p=0.5)$, $\sigma^2=1$.}
    \label{fig:latent-0.5}
\end{figure}

The next sets of plots (Figures \ref{fig:barbell-20}, \ref{fig:tree-2}, \ref{fig:tree-4},
 \ref{fig:PL-3}, \ref{fig:latent-0} and \ref{fig:latent-0.5}) highlight the bias-variance trade-off across different types of topologies. Overall, we observe a slower decay of the variance for the GraphSAGE convolution compared to the GCN convolution, for similar levels of bias. 

\begin{figure*}[!ht]
    \centering
\includegraphics[width=\linewidth]{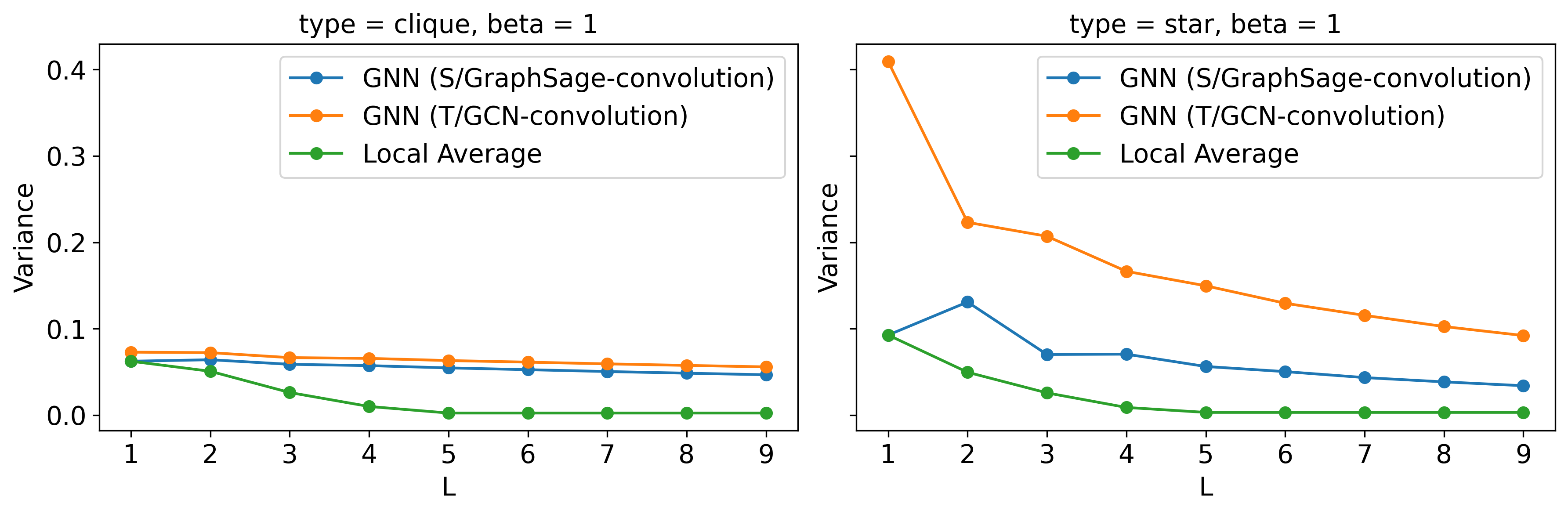}
    \caption{Variance decay of different estimators on a tree with a branching factor of 4 over four levels, with a clique of 10 nodes added (left) or a star of 10 nodes added (right).}
    \label{fig:star}
\end{figure*}
\end{document}